\documentclass[lettersize,journal]{IEEEtran}
\usepackage{amsmath,amsfonts}
\usepackage{algorithm}
\usepackage{array}
\usepackage[caption=false,font=normalsize,labelfont=sf,textfont=sf]{subfig}
\usepackage{textcomp}
\usepackage{stfloats}
\usepackage{url}
\usepackage{verbatim}
\usepackage{graphicx}
\usepackage{cite}

\usepackage{algpseudocode}    
\usepackage{color}
\usepackage[table]{xcolor}

\usepackage{booktabs}
\usepackage{multirow}
\usepackage{makecell}
\usepackage{rotating}

\begin{document}

\title{Learning Localization-aware Target Confidence for Siamese Visual Tracking}

\author{Jiahao Nie,~\IEEEmembership{Graduate Student Member,~IEEE}, Han Wu,~\IEEEmembership{Student Member,~IEEE}, Zhiwei He,~\IEEEmembership{Member,~IEEE}, Yuxiang Yang, Mingyu Gao, Zhekang Dong,~\IEEEmembership{Member,~IEEE}
\thanks{This paper was supported by the National Natural Science Foundation of China (No.61571394, No.62001149), the Key R$\&$D Program of Zhejiang Province (No.2020C03098). (\textit{Corresponding author: Zhiwei He})}
\thanks{Jiahao Nie, Han Wu, Zhiwei He and Mingyu Gao are with the School of Electronics Information, Hangzhou Dianzi University, Hangzhou 310018, China (e-mail: jhnie@hdu.edu.cn; wuhan0326@hdu.edu.cn; zwhe@hdu.edu.cn; mackgao@hdu.edu.cn)}
\thanks{Yuxiang Yang is with the School of Information Science and Technology, University of Science and Technology of China, Hefei, 230052, China (e-mail: yyx@hdu.edu.cn)}
\thanks{Zhekang Dong is with the School of Electrical Engineering, Zhejiang University, Hangzhou, 310058, China (e-mail: englishp@126.com)}}



\maketitle

\begin{abstract}
Siamese tracking paradigm has achieved great success, providing effective appearance discrimination and size estimation by the classification and regression. While such a paradigm typically optimizes the classification and regression independently, leading to task misalignment (accurate prediction boxes have no high target confidence scores). In this paper, to alleviate this misalignment, we propose a novel tracking paradigm, called SiamLA. Within this paradigm, a series of simple, yet effective localization-aware components are introduced, to generate localization-aware target confidence scores. Specifically, with the proposed \textit{localization-aware dynamic label} (LADL) loss and \textit{localization-aware label smoothing} (LALS) strategy, collaborative optimization between the classification and regression is achieved, enabling classification scores to be aware of location state, not just appearance similarity. Besides, we propose a separate localization branch, centered on a \textit{localization-aware feature aggregation} (LAFA) module, to produce location quality scores to further modify the classification scores. Consequently, the resulting target confidence scores, are more discriminative for the location state, allowing accurate prediction boxes tend to be predicted as high scores. Extensive experiments are conducted on six challenging benchmarks, including GOT-10k, TrackingNet, LaSOT, TNL2K, OTB100 and VOT2018. Our SiamLA achieves state-of-the-art performance in terms of both accuracy and efficiency. Furthermore, a stability analysis reveals that our tracking paradigm is relatively stable, implying the paradigm is potential to real-world applications.
\end{abstract}


\section{Introduction}
\IEEEPARstart{V}{isual} object tracking (VOT) is one of the fundamental tasks in computer vision, and has been widely applied to human computer interfaces, autonomous driving and smart surveillance \cite{app,things}. Given an arbitrary target object in initial frame, trackers need tell us “where” this object will appear in subsequence frames \cite{survey}. Although great progress achieved, VOT is still a challenging task to tackle, when faced with complex scenes, including occlusion, deformation, \textit{etc} \cite{vot2020, vot2021, otb2013, uav123, nfs}.

Recently, Siamese tracking paradigm \cite{siamfc,siamrpn,siamcar,ocean,siamfc++,siamban} has attracted great attention from researches, due to its well-balanced for accuracy and efficiency. The Siamese architecture consists of two streams, template and search streams. The template stream is used to extract features of the given target object in initial frame, and the search stream for feature extraction of subsequence frames. After feature extraction, a classification or regression structure follows. Early Siamese paradigm \cite{siamfc} contains only a classification branch, and the target size estimation is performed using traditional multi-scale strategy. As deep learning techniques become mature \cite{deeplearning1,deeplearning2}, the modern Siamese paradigm can be classified into two categories, anchor based \cite{siamrpn} and anchor free \cite{siamcar} paradigms. Both types contain classification and regression branches, which jointly perform the tracking task. The classification branch discriminates the appearance similarity based on the anchor or point features, and outputs the target confidence scores. Meanwhile, the predicted box with the highest target confidence score, is obtained from the regression branch.

\begin{figure*}[!t]
  \centering
  \includegraphics[width=7in]{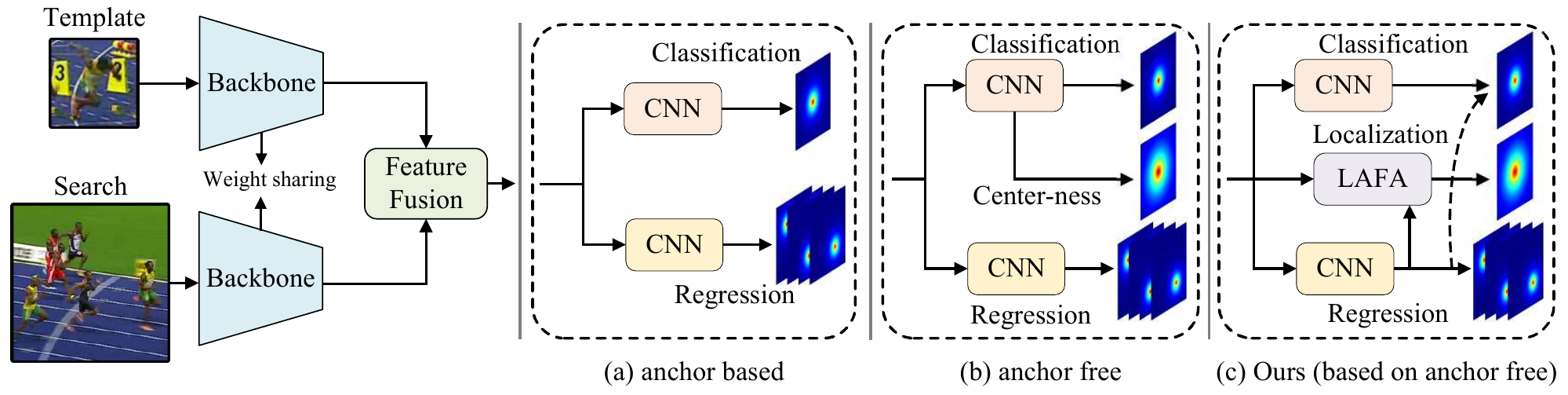}
  \caption{Comparison of our tracking paradigm (c) and the existing anchor based (a) and anchor free (b) tracking paradigms. Anchor based and anchor free paradigms are two popular pipelines in the tracking community.}
  \label{tracking frameworks}
\end{figure*}

\begin{figure}[!t]
  \centering
  \includegraphics[width=3.5in]{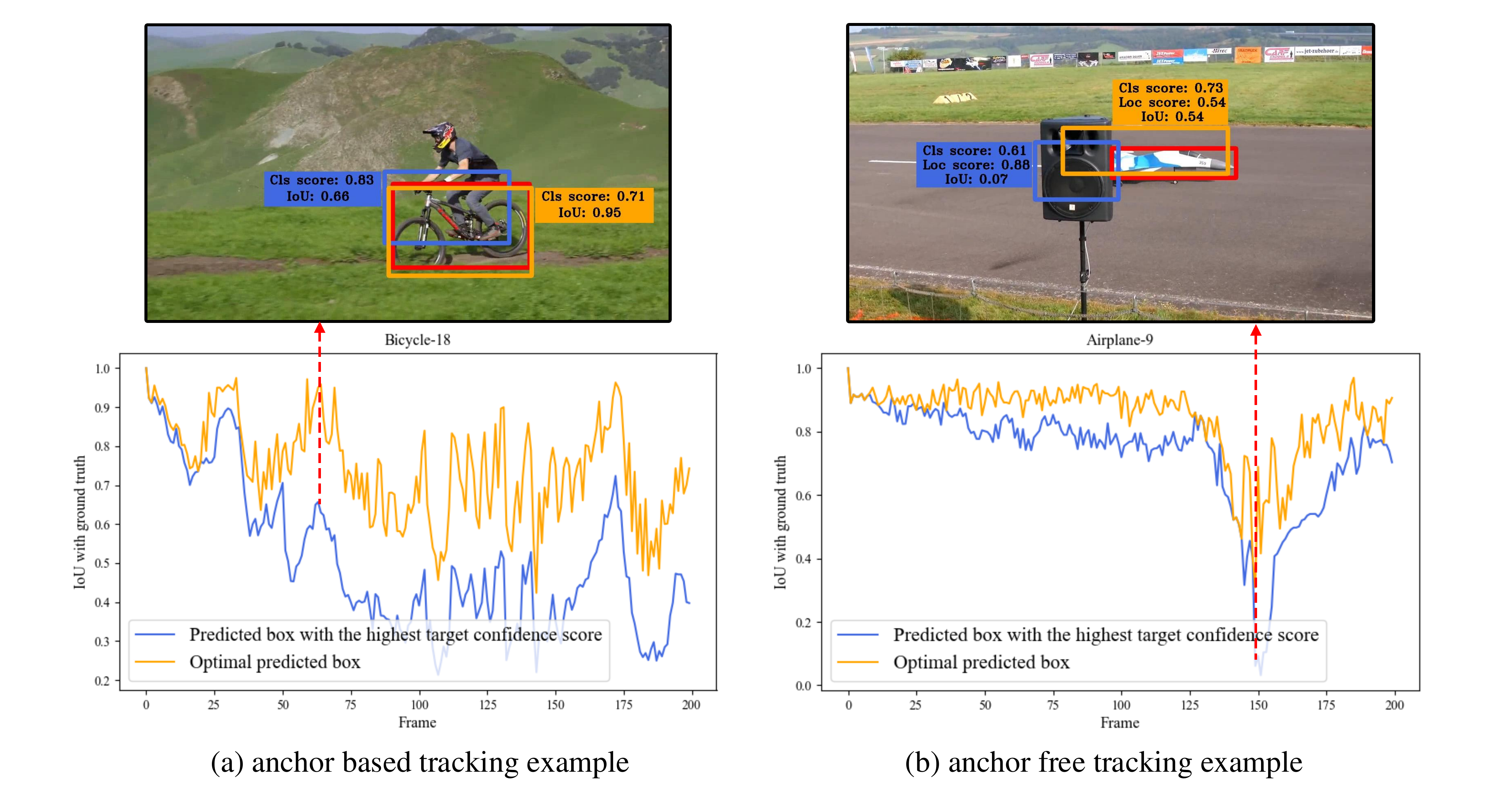}
  \caption{Specific manifestations of the task misalignment. On the left: \textbf{anchor based tracking example}, Orange and Blue lines represent the optimal prediction boxes and the tracking boxes corresponding to the highest target confidence scores. However, the optimal prediction boxes are often not selected as the tracking boxes because they usually have no highest scores (i.e., Cls score). On the right: \textbf{anchor free tracking example}, the Center-ness branch lacks localization-aware information, resulting in a large location quality score (i.e., Loc score) for the distractor.}
  \label{task misalignment}
\end{figure}

Although the development, however, such modern Siamese paradigm generally optimizes the classification and regression independently in training, failing to make them synchronized, and thus leading to the task-misalignment problem. The existing anchor based paradigm \cite{siamrpn,dasiamrpn,siamrpn++} (Fig. \ref{tracking frameworks} (a)) has rarely investigated this misalignment, so often the tracking performance falls into a suboptimal state. As illustrated in Fig. \ref{task misalignment} (a), due to the task misalignment, the predicted box with high Intersection over Union (IoU) may not have high classification score, and the contribution of the classification is greater, because the optimal box given by the regression usually has a high IoU. As a more elegant structure, the anchor free paradigm \cite{siamcar,siamfc++} (Fig. \ref{tracking frameworks} (b)) attempts to address the task misalignment via a localization branch (i.e., Center-ness). Intuitively, Center-ness predicts the potential target centers, not the location accuracy. In addition, the output location quality scores are derived from the classification branch, which lack localization-aware information, resulting in the distractor being mistaken for the target object, as shown in Fig. \ref{task misalignment} (b). Therefore, the task misalignment problem is still not well resolved.

Inspired by the above analysis, in this paper, we aim to alleviate the task misalignment, and mine the optimal prediction box as tracking box. To achieve this goal, we propose a novel Siamese tracking paradigm based on the anchor free paradigm, as illustrated in Fig. \ref{tracking frameworks} (c). The key innovation lies in the prediction head, and a series of localization-aware components are developed, to generate localization-aware target confidence scores. \textbf{(i)} Since the predicted box with the highest IoU may not have the highest classification score, we propose a \textit{localization-aware dynamic label} (LADL) loss function to alleviate this situation. Benefit from this function, the classification and regression can be optimized collaboratively, leading to accurate prediction boxes being predicted as high scores. In detail, the location state information (i.e., IoUs) from the regression branch, is used to assign labels for classification samples, allowing the classification network to discriminate positive samples in a localization-aware manner, while not treating them equally. To facilitate the collaborative optimization, we further elaborate a \textit{localization-aware label smoothing} (LALS) strategy, which increases the probability of the optimal box being involved. This is quite effective, while not presented in the existing anchor free trackers. \cite{siamcar,ocean,siamfc++,siamban} \textbf{(ii)} Due to the anchor free tracking paradigm as the baseline, we exclusively design a separate localization branch to replace the Center-ness, to further alleviate the task misalignment. Our idea is to link this branch with the regression branch to model the localization-aware information, and then deliver the location quality scores. In practice, similar to the classification training, by using IoUs from the regression branch as labels, we integrate the localization-aware information into the separate localization branch. Thanks to the separate structure, the learned features are dedicated, which fully embed localization-aware information into the localization network. When testing, the location quality scores and classification scores are combined to determine the target confidence scores, thus implicitly collaborating the classification and regression, and guiding more accurate target confidence scores for the predicted boxes. To improve the accuracy of the location quality scores, a \textit{localization-aware feature aggregation} (LAFA) module is created, which plays a significant role in the separate localization branch. Within the LAFA, we first develop a feature aggregation block, to gather point-set features to represent the predicted boxes, rather than a single point feature, which lacks significant location state information. Moreover, considering that the boundary points have poor representation for target object, we then propose a localization-aware non-local block, to capture the long- and short-term visual dependencies associated with target object, enhancing feature representation of the boundary points.

The main contributions of this work can be summarized as follows:

\begin{itemize}
  \item {We propose a novel Siamese tracking paradigm, called SiamLA. The paradigm alleviates the task misalignment in a simple yet effective, efficient and stable way.}
  \item {Following this paradigm, a series of localization-aware components are proposed, collaborating the classification and regression in the training and inference process.}
  \item {Our SiamLA outperforms the prevalent anchor based and anchor free paradigms, achieving the state-of-the-art performance on several challenging benchmarks, while running at a real-time speed of 29 fps.}
\end{itemize}

\section{Related Work}
\subsection{Siamese network based framework in visual tracking}
SiamFC \cite{siamfc} is a seminal Siamese tracking algorithm proposed by Bertinetto \textit{et al}. Due to its well-balanced for performance and efficiency, many excellent works \cite{cfnet,sasiam,structsiam,siamdw,dp} have emerged. E.g., CFNet \cite{cfnet} integrates a correlation filter into Siamese network structure, guiding a discriminative feature representation for target objects. SA-Siam \cite{sasiam} builds a twofold Siamese network to construct rich appearance and semantic information. StructSiam \cite{structsiam} extracts local structure information to identify discriminative regions of target objects. SiamDW \cite{siamdw} proposes a backbone that is more appropriate for tracking task. Although achieving promising results, these methods rarely devote to the target size estimation, using only traditional multi-scale strategy. As deep learning techniques developed, and given the relevance of object detection and object tracking tasks, regional proposal network (RPN) \cite{fastrcnn} is introduced to estimate target size in tracking, which makes the Siamese paradigm shine again. Unlike the SiamFC-series trackers, SiamRPN \cite{siamrpn} decomposes the tracking task into independent classification and regression. Benefit from the ability of the regression branch to model target shape changes, SiamRPN improves tracking accuracy, while remaining highly efficient. Based on this strong baseline, DaSiamRPN \cite{dasiamrpn} enhances the anti-interference capability by sampling more hard negative samples. SiamRPN++ \cite{siamrpn++} replaces AlexNet \cite{alxenet} with ResNet \cite{resnet} and uses layer-wise aggregation strategy, also improving tracking performance. In addition, many methods \cite{pgnet,siamacm,transt,siamgat,learntomatch,ttdimp,multi-level,bt} have investigated the feature fusion part, i.e., the neck of a model. After this anchor based tracking period, anchor free tracking paradigm has become prevalent recently. Inspired by the anchor free mechanism in detection, many anchor free trackers \cite{siamcar, ocean, siamfc++, siamban,siamcorners,cat} have been designed. Compared with the anchor based structure, anchor free design discards the tracker-sensitive anchors hyper-parameter, and thus leading to a more elegant and efficient tracking paradigm. Besides, an additional localization branch is employed in the anchor free paradigm, to predict the location quality of the predicted boxes. In this paper, the proposed Siamese tracking paradigm is based on the anchor free paradigm, and aims to alleviate the task misalignment mentioned in section 1.
\subsection{task misalignment in visual tracking}
Generally, the modern Siamese trackers rely on the target confidence scores as a criterion for tracking. But most of them optimize the classification and regression independently in training, inevitably resulting in that the predicted boxes with high target confidence scores may not accurate. Towards this issue, some multi-stage methods \cite{spm,c-rpn,multi-stage} have been proposed. SPM \cite{spm} employs SiamRPN as the first stage to generate high quality proposals, followed by a lightweight network as the second stage to select the optimal one. C-RPN \cite{c-rpn} iterates continuously to produce more accurate prediction boxes by a cascaded RPN. These multi-stage paradigm trackers are trained stage by stage and operate inefficiently. As a better solution, SiamRCR \cite{siamrcr} implements task-alignment by re-weight the losses for each positive sample, allowing the model to pay more attention to high-quality samples with high IoUs. In contrast, we propose a \textit{localization-aware dynamic label} (LADL) loss, making the classification and regression branches to be optimized collaboratively. In particular, we elaborate a simple, yet effective sampling strategy, \textit{localization-aware label smoothing} (LALS). It promotes the collaborative optimization, further increasing the probability of the optimal box being involved. To tackle the task misalignment better, we creatively propose a separate localization branch to implicitly collaborate the classification and regression. The work related to the localization branch is presented below.
\subsection{Localization branch in visual tracking}
Localization branch serves as a unique structure in the anchor free paradigm. SiamFC++ \cite{siamfc++} and SiamCAR \cite{siamcar} use Center-ness \cite{fcos} to assess the location quality, and deliver the location quality scores to modify the classification scores. Nevertheless, the Center-ness fails to address the task misalignment problem due to the lack of localization-aware information. Additionally, in the post-processing process, the penalty window plays a similar role to the Center-ness, which downplays the role of the Center-ness. To leverage localization branch to alleviate the task misalignment, we abandon the Center-ness design philosophy, and separate this branch, of which a \textit{localization-aware feature aggregation} (LAFA) module is proposed as its core.

\section{Proposed Method}
\begin{figure*}[!t]
  \centering
  \includegraphics[width=7in]{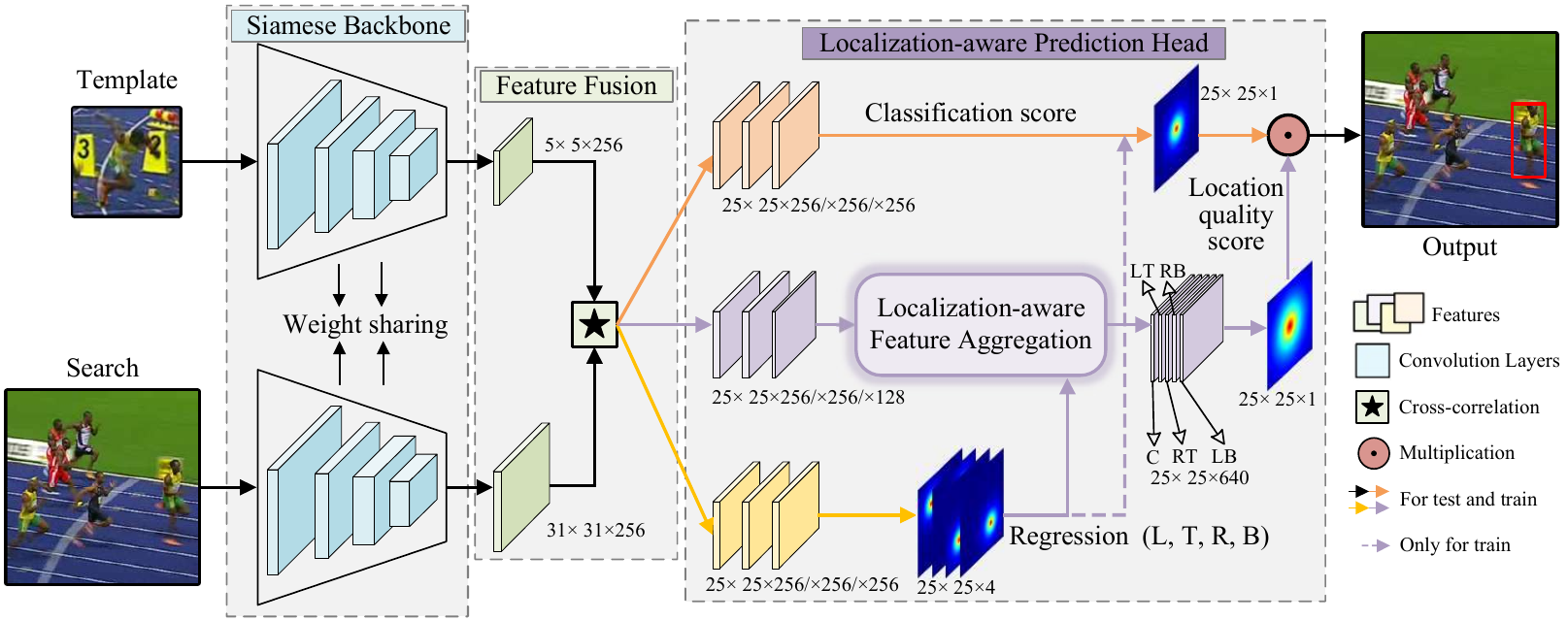}
  \caption{An overview of our Siamese tracking paradigm SiamLA, consisting of a Siamese backbone, a feature fusion module and a localization-aware prediction head. Thanks to the proposed \textit{localization-aware dynamic label} (LADL) and \textit{localization-aware label smoothing} (LALS), the classification and regression can be optimized collaboratively. In addition, a separate localization branch, with a \textit{localization-aware feature aggregation} (LAFA) module as its core, is designed to implicitly collaborate the classification and regression.}
  \label{siamla}
\end{figure*}
In this work, we propose a novel Siamese tracking paradigm, called SiamLA, as shown in Fig. \ref{siamla}. The implementation procedure is presented in this section. We briefly revisit the baseline tracker with the testing and training phases in section 3.1. Then, we detail the proposed LADL, LALS and LAFA in section 3.2, 3.3 and 3.4, respectively.
\subsection{Revisit of the baseline method}
Here, we employ a classic anchor free tracker SiamCAR \cite{siamcar} as our baseline. The testing and training process are as follows. More details of testing and training can be found in \cite{siamcar}.

\textbf{Testing: }As illustrated in Fig. \ref{siamla}, the template image $z$ and search image $x$ are fed into the Siamese backbone, to extract the features $\phi(z)$ and $\phi(x)$. Then, a cross-correlation operator is applied between the features $\phi(z)$ and $\phi(x)$ to fuse them, which transfers the template information to the search region for subsequent tasks, classification and regression. The classification scores map $c_{i,j}$, regression offset map $r_{i,j}$ and location quality scores map $C_{i,j}$ are calculated by
\begin{equation}
  \begin{aligned}
    \{c_{i,j}\} &={{f}_{cls}}(Cr{{b}_{cls}}(\phi (z)\star \phi (x))), \\ 
    \{r_{i,j}\} &={{f}_{reg}}(Cr{{b}_{reg}}(\phi (z)\star \phi (x))), \\
    \{C_{i,j}\} &={{f}_{cen}}(Cr{{b}_{cls}}(\phi (z)\star \phi (x)))   
  \end{aligned}
  \label{eq1}
\end{equation}
where $\star$ denotes the cross-correlation operator, $\phi$ represents the Siamese backbone, $Crb$ is a series of “Conv-Relu-Bn”s, and the convolution layer $f_{cls/cen/reg}(.)$ outputs the corresponding map. Finally, the predicted box corresponding with the highest target confidence score (i.e., the classification score multiplied by the location quality score) will be considered as the tracking result.

\textbf{Training: }SiamCAR is trained with three losses. Cross-entropy loss $L_{ce}$, IoU loss $L_{iou}$ and binary cross-entropy loss $L_{bce}$ are used to optimize the classification, regression and Center-ness branches, respectively, as follows
\begin{equation}
  \begin{aligned}
   {{\mathcal{L}}_{cls}}&=\frac{1}{{{N}_{cls}}}\sum_{i,j}{{{L}_{ce}}(c_{i,j},\hat{c}_{i,j})}, \\ 
   {{\mathcal{L}}_{reg}}&=\frac{1}{{{N}_{reg}}}\sum_{i,j}{\mathbb{I}_{\{\hat{c}_{i,j}=1\}} {{L}_{iou}}(r_{i,j},\hat{r}_{i,j})}, \\ 
   {{\mathcal{L}}_{cen}}&=\frac{1}{{{N}_{cen}}}\sum_{i,j}{\mathbb{I}_{\{\hat{c}_{i,j}=1\}} {{L}_{bce}}(C_{i,j},\hat{C}_{i,j})} 
  \end{aligned}  
  \label{eq2}
\end{equation}
where $\hat{c}_{i,j}$, $\hat{r}_{i,j}$ and $\hat{C}_{i,j}$ represent the corresponding labels. For the classification, pixels within a certain size rectangle, centered on the target object, are defined as positive samples, and the other pixels of the image are defined as negative samples. For the regression and Center-ness, only positive samples are sampled for training.

\subsection{Localization-aware dynamic label}
In the existing anchor free trackers, the classification and regression independently optimize their respective loss objectives in training, resulting in the task misalignment, i.e., the prediction boxes corresponding to points with high target confidence scores may not accurate. In practice, however, we expect the higher the target confidence score is, the more accurate the corresponding prediction box will be. To satisfy this, the parallelized training style is discarded. The IoUs of the predicted boxes in the regression branch, are adopted as dynamic labels for the classification samples. Here, we redefine the training objective for classification, this novel \textit{localization-aware dynamic label} loss can be formulated as
\begin{equation}
  {{\mathcal{L}}_{ladl}}=\frac{1}{{{N}_{ladl}}}\sum_{i,j}{{\rm{LADL}}(c_{i,j},\hat{c}_{i,j})}
  \label{eq3.1}
\end{equation}
where the $\rm{LADL}$ loss function as follow
\begin{equation}
  {\rm{LADL}}(c_{i,j},\hat{c}_{i,j}) =
  \begin{cases}
  -(\hat{IoU_{i,j}}{\rm{log}}(1-c_{i,j})+\\
  (1-\hat{IoU_{i,j}}){\rm{log}}(1-c_{i,j})) & \text{if } \hat{c}_{i,j} = 1,\\
  -{\rm{log}}(1-c_{i,j})(1+c_{i,j}) & \text{if } \hat{c}_{i,j} = 0
  \end{cases} 
  \label{eq3}
\end{equation}
where the predicted classification scores map $c\in{\mathbb{R}^{H \times W\times 1}}$, and the $\hat{IoU_{i,j}}$ is calculated by
\begin{equation}
  \hat{IoU_{i,j}}=\frac{1}{1+{{e}^{(IoU_{i,j}-\alpha )\times \beta}}}    
  \label{eq4}
\end{equation}
where $IoU_{i,j}\in{[0,1]}$ and the parameters $\alpha$ and $\beta$ are set to 0.5 and 10 to guide the output $\hat{IoU_{i,j}}\in{(0,1)}$. Equation \ref{eq4} reduces the output variation at both ends, making the classification easier to be optimized. Moreover, the low output variation at both ends, also increases the stability, because the ground truth is annotated with reasonable noise. 

Notably, as illustrated in Equation \ref{eq3}, the classification scores $c_{i,j}$ dynamically re-weight the negative samples, enabling the classification network to focus more on hard negative samples with high classification scores. During the training, considering that no accurate prediction box is available in the initial stage, we normalize the dynamic labels at image-level by
\begin{equation}
  \hat{IoU_{i,j}}=\frac{\hat{IoU_{i,j}}-min(\{\hat{IoU_{i,j}}\})}{max(\{\hat{IoU_{i,j}}\})-min(\{\hat{IoU_{i,j}}\})}    
  \label{eq5}
\end{equation}
where the output $\hat{IoU_{i,j}}\in{[0,1]}$.

\subsection{Localization-aware label smoothing}
\begin{figure}[!t]
  \centering
  \includegraphics[width=1.6in]{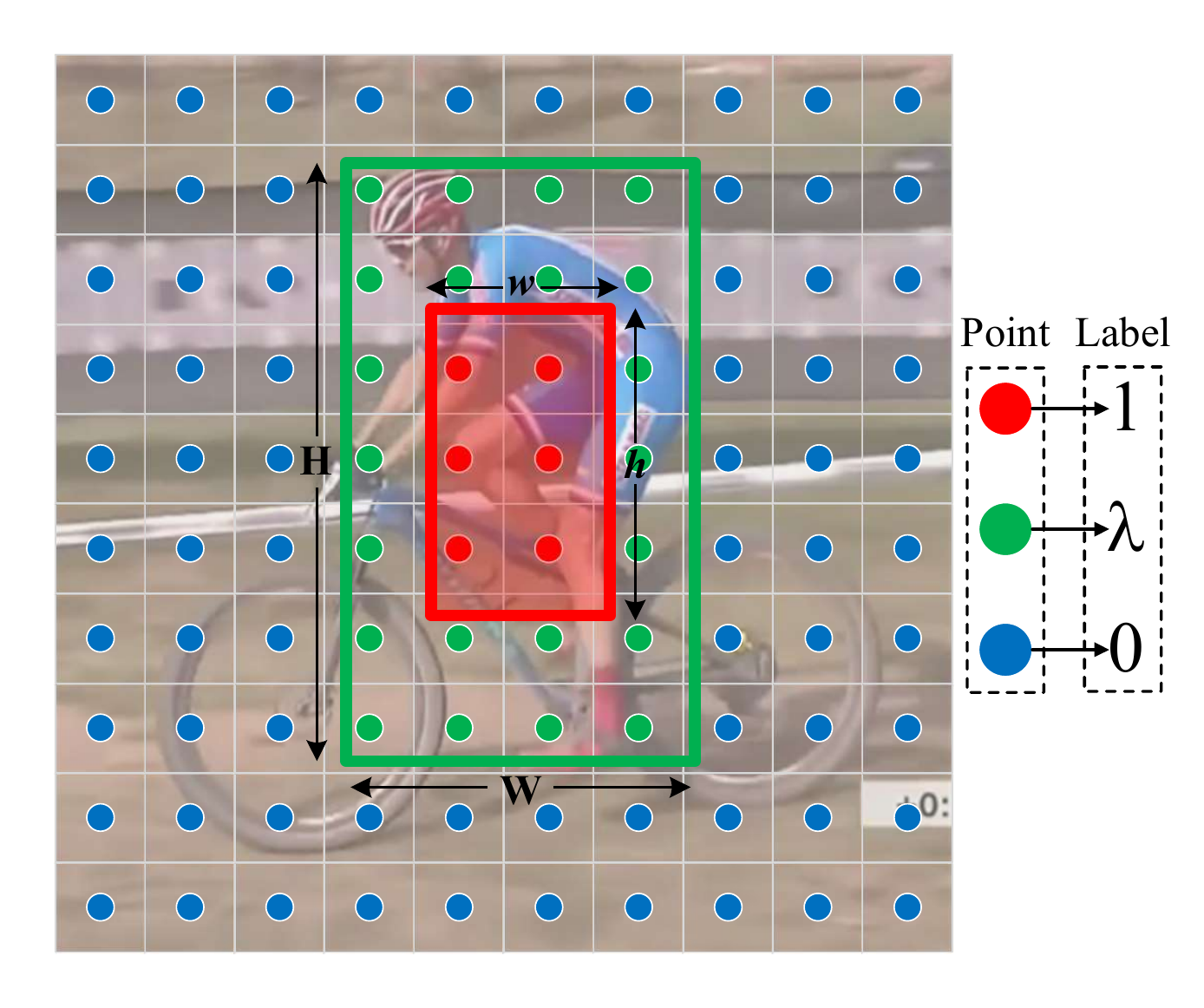}
  \includegraphics[width=1.6in]{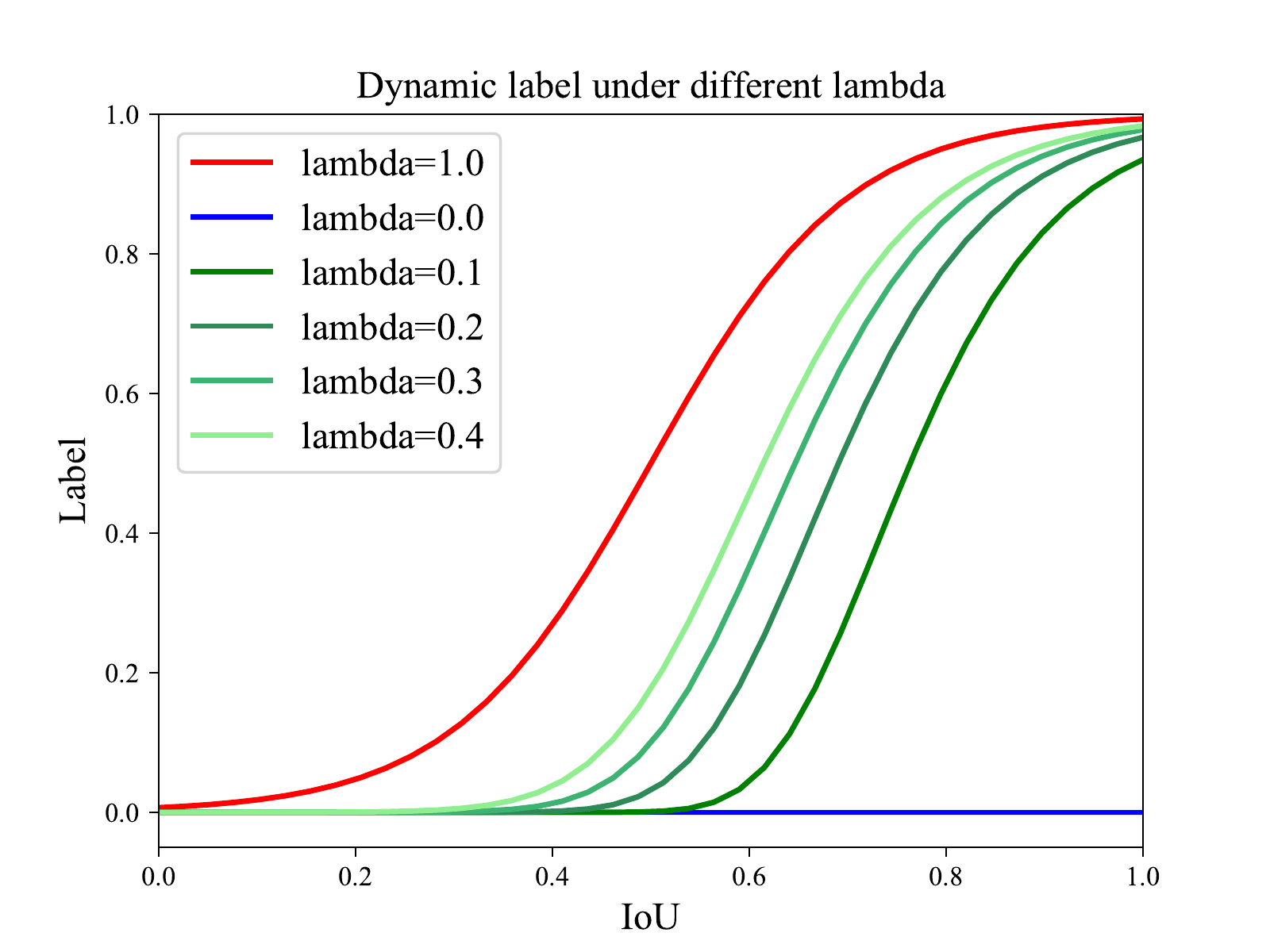}
  \caption{Illustration of the \textit{localization-aware label smoothing} strategy. \textbf{On the left}, the initial labels for samples are defined. \textbf{On the right}, we smooth the initial labels using IoUs and a hyper-parameter $\lambda$. $\lambda=1$ represents the red points (positive samples), $\lambda=0$ represents the blue points (negative samples). In terms of the green points (regularized positive samples), the smoothing curves at different $\lambda$ are presented.}
  \label{lambda}
\end{figure}

Although the above \textit{localization-aware dynamic label} effectively alleviate the task misalignment, it still fail to mine the optimal prediction box, which may be regressed by the boundary points. This is due to that the boundary points are sampled as negative samples to train classification network. To solve this problem, we elaborate a \textit{localization-aware label smoothing} strategy to define the labels of the boundary points. As illustrated in Fig. \ref{lambda} (a), we consider the boundary (green points) as positive samples as well, and smooth their labels as $\lambda$. In fact, the target confidence scores of the center points, are expected to be higher than the boundary points, which reduces the occurrence of tracker drift. Therefore, we make a distinction between positive samples at the center and boundary points. In detail, for the center samples (red points), the labels are completely determined by the corresponding IoUs, while for the boundary samples (green points), a positive sample is considered only if the IoU is large enough, and its label should be smaller than the center sample with the same IoU. To implement this, we reformulate the Equation \ref{eq4} as
\begin{equation}
  \hat{IoU_{i,j}}=(\frac{1}{1+{{e}^{(IoU_{i,j}-\alpha )\times \beta}}})^{\frac{1}{\lambda}}     
  \label{eq6}
\end{equation}
where $\lambda=1$ for center samples, other $\lambda$ for boundary samples. Fig. \ref{lambda} (b) shows the label changes under different $\lambda$, it is evident that the Equation \ref{eq6} satisfies our requirements. The parameter $\lambda$ is explicitly discussed in section 4.

\subsection{Localization-aware voting mechanism}
Most of the existing anchor free trackers use Center-ness as localization branch, where is induced from the classification branch. However, such a structure lacks localization-aware information, leaving the task misalignment problem unresolved. Therefore, we separate this branch, and use the localization-aware information from the regression branch to supervise its training. Besides, we also propose a \textit{localization-aware feature aggregation} module that contains a localization-aware non-local block and a feature aggregation block, guiding more accurate location quality scores.
\begin{figure}[!h]
  \centering
  \includegraphics[width=3.5in]{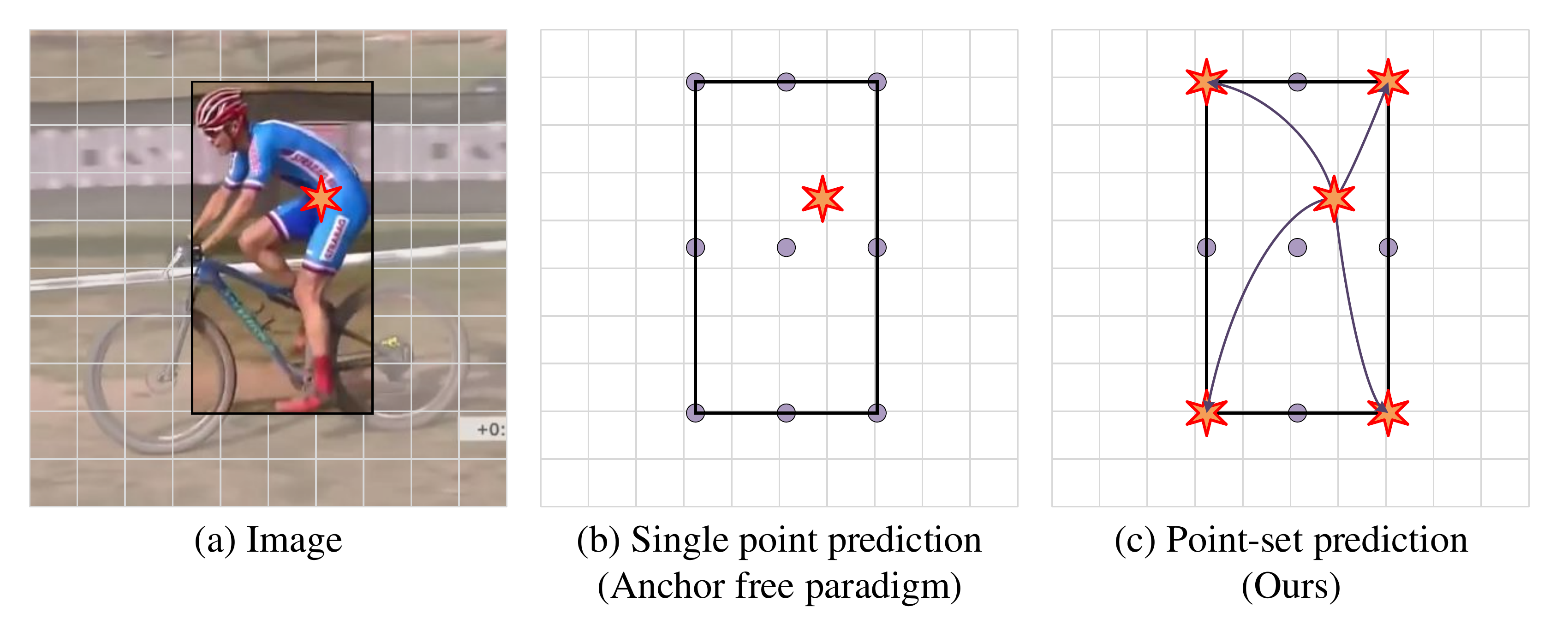}
  \caption{Comparison between the prediction process of our feature aggregation block and normal anchor free paradigm.}
  \label{feature aggregation}
\end{figure}
\begin{figure}[!t]
  \centering
  \includegraphics[width=3in]{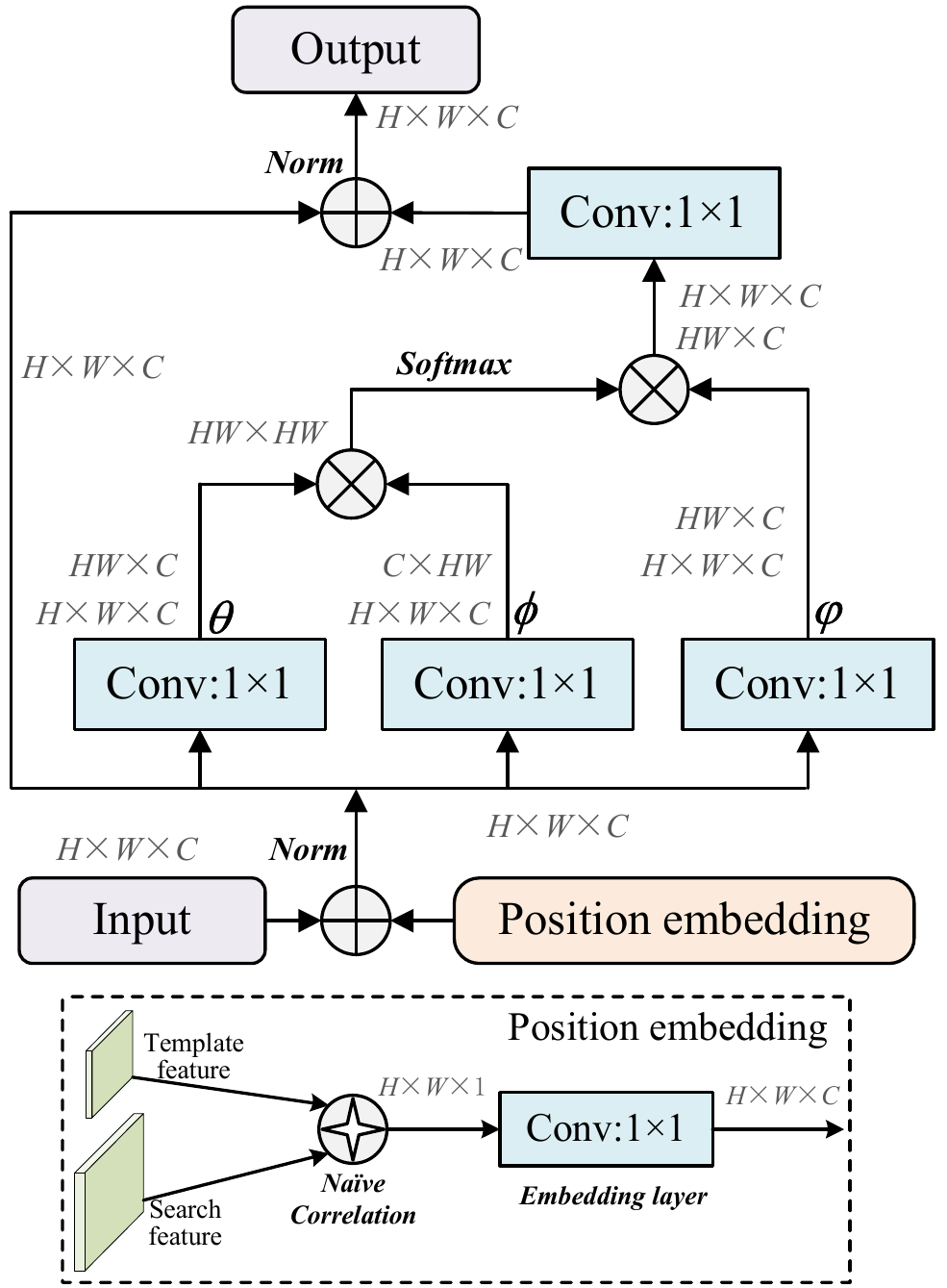}
  \caption{An overview of the proposed localization-aware non-local block.}
  \label{localization-aware non-local}
\end{figure}

In practice, as show in Fig. \ref{feature aggregation}, we propose a feature aggregation block, to aggregate the features of the center, top-left, top-right, bottom-right, and bottom-left points of the predicted boxes, instead of relying only on the single center point features. Nevertheless, one shortcoming exists in this block is that the boundary point feature has a week representation for target objects. Thus, we develop an efficient localization-aware non-local block, to capture the long- and short-term visual dependencies associated with the target objects. In contrast to the normal non-local structure, we incorporate a learnable position embedding, making this block be aware of location state. This is inspired by TrDiMP \cite{trdimp} and ToMP \cite{tomp}, we introduce positional prior knowledge to generate the position embedding. But differently, we employ a naïve correlation operator $h(.)$ between the template and search features, to highlight the target position, while not using a labeled mask presented in \cite{trdimp, tomp}. A $1\times 1$ convolution layer $e(.)$ then is used to expand the channels as follow
\begin{equation}
  p=e(h(\phi(z),\phi(x)))     
  \label{eq7}
\end{equation}
The data flow of the whole localization-aware non-local block is depicted in Fig. \ref{localization-aware non-local}, which can be expressed using an equation as
\begin{equation}
  y=f(Softmax(\theta(x+p)\phi^{\top}(x+p))\varphi(x))+x     
  \label{eq8}
\end{equation}
where $f(.)$, $\theta(.)$, $\phi(.)$ and $\varphi(.)$ denote a $1\times 1$ convolution layer, $x$ and $p$ represent the input features and the introduced position embedding, respectively.

For training the separate localization branch, similar to Equation \ref{eq3}, we use IoUs as supervised information. Besides, we sample the negative samples in a 1:1 (positive: negative) ratio to deepen the anti-distractors capability of the localization network. The optimization objective is formulated as
\begin{equation}
  \mathcal{L}_{lavm}=\frac{-1}{{{N}_{lavm}}}\sum_{i,j}{{{L}_{bce}}(c_{i,j},IoU_{i,j})}     
  \label{eq9}
\end{equation}
For testing, the location quality scores serve as an effective constraint on the classification scores, to guide the more accurate target confidence scores. This implicitly collaborates the classification and regression, thus also alleviating the task misalignment to some extent.

\section{Experiments}
\subsection{Implementation details}
\textbf{Training details: }The modified ResNet-50 \cite{resnet}, that is initialized with the parameters pre-trained on ImageNet \cite{imagenet}, is employed as the backbone network of the proposed tracker SiamLA. Following the setting of the anchor free trackers \cite{siamfc++,siamban,siamcar,ocean}, we use the training splits of the Youtube-BB \cite{ytbb}, ImageNet VID \cite{imagenet}, ImageNet DET \cite{imagenet}, GOT-10k \cite{got10k} and COCO \cite{coco} datasets to train the model offline. The template and search images are cropped to 127 pixels and 255 pixels centered on the target objects. The overall training objective can be defined as
\begin{equation}
  \mathcal{L}={{\mathcal{L}}_{ladl}}+{{\lambda }_{1}}{{\mathcal{L}}_{reg}}+{{\lambda }_{2}}{{\mathcal{L}}_{lavm}}
  \label{eq10}
\end{equation}
where $\lambda_1$ and $\lambda_2$ are weights, both set to 1 to balance the different training losses. A SGD optimizer is utilized to optimize this overall training objective, and the batch size is set to 128 images. With a total of 50 epochs trained, for the first 5 epochs, we freeze the backbone and train the localization-aware prediction head. For the remaining epochs, the whole model is trained end-to-end with an exponentially decay of the learning rate from $5\times 10^{-3}$ to $10^{-5}$. Our code is implemented using PyTorch 1.8.0 and Python 3.6. The experiments are conducted on a workstation with an Intel i7-10700F @ 2.9 GHz CPU, and 4 NVIDIA GTX 1080Ti GPUs with CUDA10.2.

\textbf{Testing details: }For a fair comparison, we use the same testing strategy with SiamCAR \cite{siamcar}, to evaluate the proposed tracker SiamLA. Since the anchor free tracking paradigm is inherited, the tracking process is treated as a point selection by the target confidence scores. The primary target confidence scores are determined by a 2D vector $(cls_{i,j},loc_{i,j})$, in which $cls_{i,j}$ and $loc_{i,j}$ represent the outputs of the classification and localization branches, respectively. Based on the fact that the subtle target changes happened in adjacent frames, we use two penalty factors to re-rank the target confidence scores, as follow
\begin{equation}
  s_{i,j}=(1-w)cls_{i,j}\times loc_{i,j}\times p_{i,j} + wW_{i,j}
  \label{eq11}
\end{equation}
where $W$ is the cosine window, weighted by a window$\_$influence weight $w$. $p_{i,j}$ is the penalty factor for scale change, is formulated by
\begin{equation}
  {{p}_{i,j}}={{e}^{k\times \max (\frac{r}{{{r}^{'}}},\frac{{{r}^{'}}}{r})\times \max (\frac{s}{{{s}^{'}}},\frac{{{s}^{'}}}{s})}}
  \label{eq12}
\end{equation}
where $k$ is a hyper-parameter, $r$ and $r^{'}$ represent the aspect ratio $\frac{h}{w}$ of the predicted boxes in the current and previous frames, $s$ and $s^{'}$ represent the sizes of the predicted boxes. The mean of the predicted boxes corresponding to the top-n target confidence scores $s_{i,j}$ is regarded as the tracking result, and the n is set to 3 as in \cite{siamcar}. The entire tracking process is presented in Algorithm \ref{al1}.
\begin{algorithm}[h]  
  \caption{Tracking with SiamLA}
  \label{al1}  
  \begin{algorithmic}[1]  
    \Require  
      Images of a video sequence ${\{I_t\}_{t=1}^{T}}$;  
      The given target box $b_1$ in initial frame $I_1$. 
    \Ensure  
    Predicted boxes $\{b_t\}_{t=2}^{T}$ in subsequence frames. \\
    Crop the template image $z$ using $b_1$ for $I_1$. \\
    Extract feature $\phi(z)$ of $z$.
    \For{$t=2$ to $T$} 
      \State Crop the search image $x$ using $b_{t-1}$ for $I_t$;  
      \State Extract feature $\phi(x)$ of $x$;
      \State Obtain the target confidence scores $\{s_{i,j}\}$ by Eq. \ref{eq12};
      \State Obtain $b_t$ using $\{s_{i,j}\}$ and the post-processing algorithm in SiamCAR \cite{siamcar};
    \EndFor 
  \end{algorithmic}  
\end{algorithm}

\textbf{Evaluation benchmarks and metrics: }We evaluate the proposed tracker SiamLA on six challenging benchmarks, including GOT-10k \cite{got10k}, TrackingNet \cite{trackingnet}, LaSOT \cite{lasot}, TNL2K \cite{tnl2k}, OTB100 \cite{otb2015} and VOT2018 \cite{vot2018}. GOT-10k uses average overlap (AO) and success rate (SR) to evaluate the performance. For the TrackingNet and LaSOT benchmarks, three metrics, i.e., area under curve (AUC), precision (Prec.), plus the normalized precision (N$\_$Prec.) are adopted. Similarly, on the TNLK2 and OTB100 benchmarks, the trackers are ranked using AUC and Prec. scores. All these benchmarks use a one-pass-evaluation (OPE) to run a tracker. For the VOT2018 benchmark, whenever tracking fails, a tracker will reinitialize a new frame as the template image. Based on this running style, two metrics accuracy and robustness are derived to evaluate the performance. In addition, the expected average overlap (EAO) is a comprehensive metric that takes both accuracy and robustness into account.

\subsection{Comparison with the state-of-the-arts}
\begin{table*}[!t]
  \caption{State-of-the-art Comparison on GOT-10k \cite{got10k}, TrackingNet \cite{trackingnet} and LaSOT \cite{lasot}. The best three results are colored by red, green and blue.}
  \label{table-gtl}
  \centering
  \begin{tabular}{cccccccccccccccc}
    \toprule
    \multicolumn{1}{c}{\multirow{2}*{Paradigm}}
    & \multicolumn{1}{c}{\multirow{2}*{Tracker}}
    &\multicolumn{3}{c}{GOT-10k \cite{got10k}} & &\multicolumn{3}{c}{TrackingNet \cite{trackingnet}} & &\multicolumn{3}{c}{LaSOT \cite{lasot}} & &\multicolumn{2}{c}{Speed}\\
    \cmidrule{3-5}  \cmidrule{7-9} \cmidrule{11-13} \cmidrule{15-16}
    & & AO & SR$_{0.5}$ & SR$_{0.75}$ && AUC & Prec. & N$\_$Prec. && AUC & Prec. & N$\_$Prec. && Device & Fps\\
    \midrule
    \multirow{5}{*}{SiamFC-series} 
    & SiamFC \cite{siamfc}& 0.392 & 0.426 & 0.135 && 0.571 & 0.533 & 0.654 && 0.336 & 0.339 & 0.420 && Titan X & 86 \\
    & CFNet \cite{cfnet}& 0.434 & 0.481 & 0.190 && 0.578 & 0.533 & 0.654 && 0.275 & 0.259 & 0.312 &&  Titan X & 52\\
    & DSiam \cite{dsiam}& 0.417 & 0.461 & 0.149 && 0.333 & - & 0.405 && 0.333 & 0.322 & 0.405 &&  Titan X & 45\\
    & StructSiam \cite{structsiam}& - & - & - && - & - & - && 0.335 & 0.337 & 0.422 &&  GTX 1080 & 45\\
    & SiamDW \cite{siamdw}& 0.411 & 0.456 & 0.154 && - & - & - && 0.384 & 0.356 & - && GTX 1080 & 45\\
    \midrule
    \multirow{8}{*}{Anchor-based}
    & SiamRPN \cite{siamrpn}& 0.444 & 0.536 & 0.222 && - & - & - && 0.447 & 0.432 & 0.542 && GTX 1060 X & 160 \\
    & DaSiamRPN \cite{dasiamrpn}& 0.481 & 0.581 & 0.270 && 0.638 & 0.591 & 0.733 && 0.515 & 0.529 & 0.605 && Titan X & 110\\
    & SiamRPN++ \cite{siamrpn++}& 0.518 & 0.618 & 0.325 && 0.733 & 0.694 & 0.800 && 0.496 & 0.491 & 0.569 && Titan Xp & 35\\
    & SiamLTR \cite{siamltr}& 0.593 & 0.698 & 0.474 && 0.736 & 0.691 & 0.802 && 0.525 & 0.533 & - && GTX 1080Ti & 35\\
    & SPM \cite{spm}& 0.513 & 0.593 & 0.359 && 0.712 & 0.660 & 0.771 && - & -& - && Tesla P100 & 120 \\
    & C-RPN \cite{c-rpn}& - & - & - && 0.669 & 0.619 & 0.746 && 0.455 & 0.425 & - && GTX 1080Ti & 36\\
    & SPLT \cite{splt}& - & - & - && - & - & - && 0.426 & 0.396 & 0.494 && GTX Titan X & 26\\
    & D3S \cite{d3s}& 0.597 & 0.676 & 0.415 && 0.728 & 0.664 & 0.768 && 0.488 & 0.490 & - && GTX 1080 & 25\\
    & ROAM \cite{roam}& 0.465 & 0.532 & 0.236 && 0.670 & 0.623 & 0.754 && 0.447 & 0.445 & - && RTX 2080 & 13\\
    \midrule
    \multirow{8}{*}{Anchor-free}
    & SiamBAN \cite{siamban}& 0.579 & 0.684 & 0.457 && -& - & - && 0.514 & 0.521 & 0.598 && GTX 1080Ti & 40 \\
    \rowcolor{black!10} & SiamCAR \cite{siamcar}& 0.581 & 0.683 & 0.441 && 0.740 & 0.684 & 0.804 && 0.516 & 0.524 & 0.610 && GTX 1080Ti & 38 \\
    & SiamFC++ \cite{siamfc++}& 0.595 & 0.695 & {\color{blue}0.497} && {\color{blue}0.754} & {\color{blue}0.705} & 0.800 && 0.544 & 0.547 & 0.623 && RTX 2080Ti & 90\\
    & Ocean \cite{ocean} & 0.611 & 0.721 & 0.473 && - & - & - && 0.560 & {\color{blue}0.566} & {\color{green}0.651} && Tesla V100 & 25 \\
    & SiamRCR \cite{siamrcr}& {\color{green}0.624} & {\color{red}0.752} & 0.460 && {\color{green}0.764} & {\color{green}0.716} & {\color{green}0.818} && {\color{red}0.575} & {\color{red}0.599} & - && Tesla P40 & 65\\
    & SiamGAT \cite{siamgat} & {\color{red}0.627} & {\color{green}0.743} & 0.488 && 0.753 & 0.698 & {\color{blue}0.807} && 0.539 & 0.530 & {\color{blue}0.633} && RTX 2080Ti & 70\\
    & SBT-light \cite{sbtlight} & 0.602 & 0.685 & {\color{red}0.530} && - & - & - && {\color{green}0.565} & {\color{green}0.571} & - && Tesla V100 & 62\\
    \midrule
    \rowcolor{black!10} 
    \multirow{1}{*}{Ours}
    & SiamLA & {\color{blue}0.619} & {\color{blue}0.724} & {\color{green}0.510}&&{\color{red}0.767} & {\color{red}0.718} & {\color{red}0.821} && {\color{blue}0.561} & 0.560 & {\color{red}0.652} && GTX 1080Ti & 29 \\
    \bottomrule
  \end{tabular}
\end{table*}
\textbf{GOT-10k: }GOT-10k \cite{got10k} is a large-scale generic object tracking benchmark, and the test set contains 180 video sequences with 84 object classes. The ground truths are not open, so the performance of a tracker needs to be obtained by submitting the results to an evaluation service platform. We follow the protocols in \cite{got10k}, using only GOT-10k train set to train our SiamLA. As shown in Table \ref{table-gtl}, the proposed tracker SiamLA achieves the state-of-the-art performance, getting 0.619, 0.724 and 0.510 scores in terms of AO, SR$_{0.5}$ and SR$_{0.75}$, respectively. Although SiamRCR and SiamGAT exhibit a slightly inferior performance in AO or SR$_{0.5}$, our SiamLA has a significant advantage in SR$_{0.75}$. Compared with the baseline SiamCAR, the SiamLA improves 3.8\% and 4.1\% in AO and SR$_{0.5}$ scores. Especially for the SR$_{0.75}$, with the IoU threshold set to 0.75, our tracking success rate improves by 6.9 points, indicating that the task misalignment is effectively alleviated, i.e., the proposed Siamese tracking paradigm is able to mine the optimal prediction boxes.
\begin{figure}[!t]
  \centering
  \includegraphics[width=2.5in]{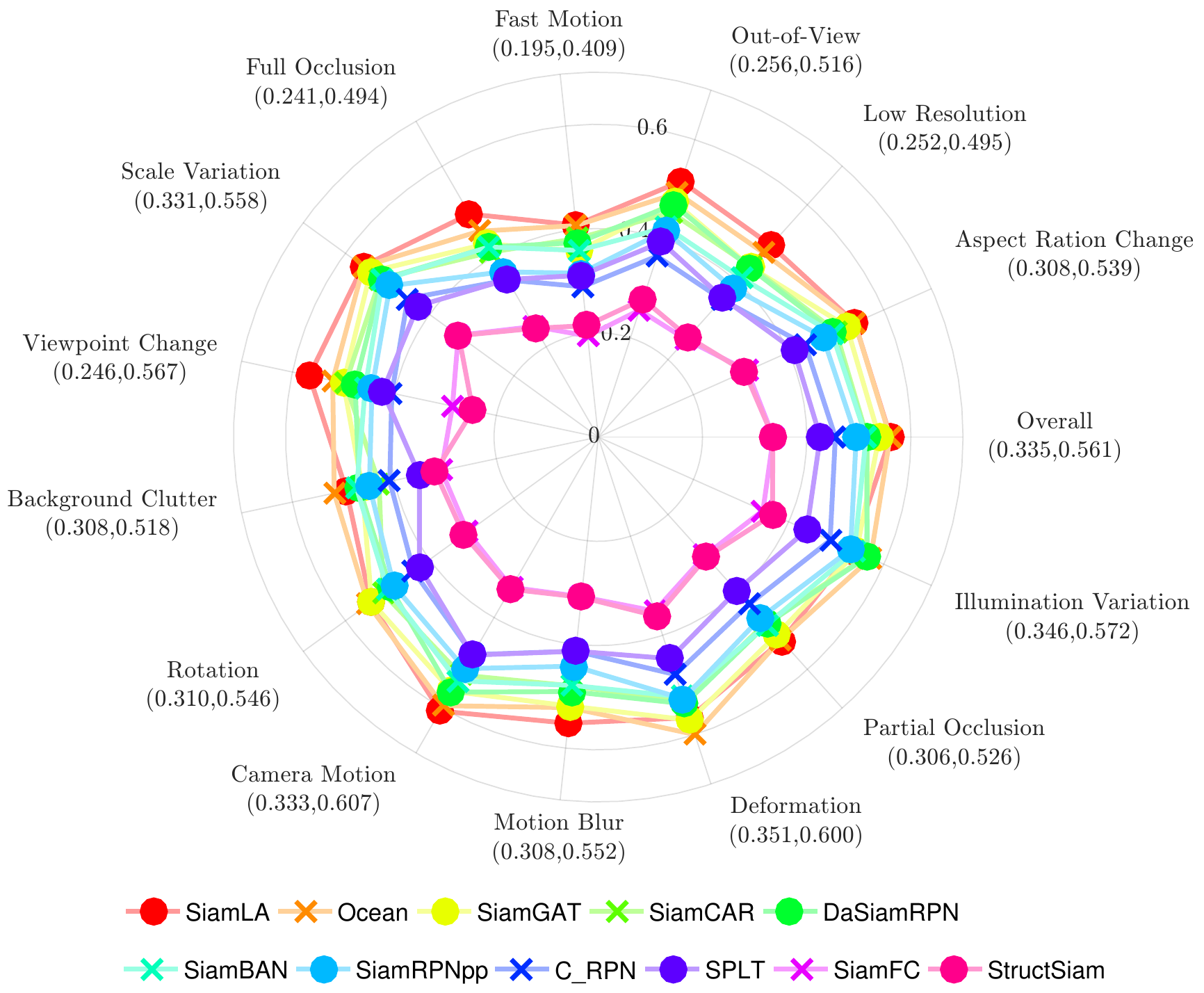}
  \caption{AUC scores of 14 attributes on the LaSOT \cite{lasot} test set.}
  \label{lasot_attr_suc}
\end{figure}

\textbf{TrackingNet: }TrackingNet \cite{trackingnet} is a large-scale short-term tracking benchmark that collected from the real world, with a test set of 511 video sequences. Similar to the GOT-10k, the tracking results also need to be evaluated on a specific service platform. The performance comparison on this benchmark is presented in Table \ref{table-gtl}. Our tracker SiamLA gets the best performance among all the comparison state-of-the-art trackers, obtaining 0.767, 0.718 and 0.821 scores in terms of AUC, Prec. and N$\_$Prec., respectively. Despite the baseline SiamCAR exhibits impressive tracking performance, our SiamLA still outperform it by 2.7, 3.4 and 1.7 points in the three metrics. Since the various wild scenes are involved in the test set, the excellent performance demonstrates the effectiveness and generality of the proposed tracking paradigm in real world.
\begin{figure}[!t]
  \centering
  \includegraphics[width=3.5in]{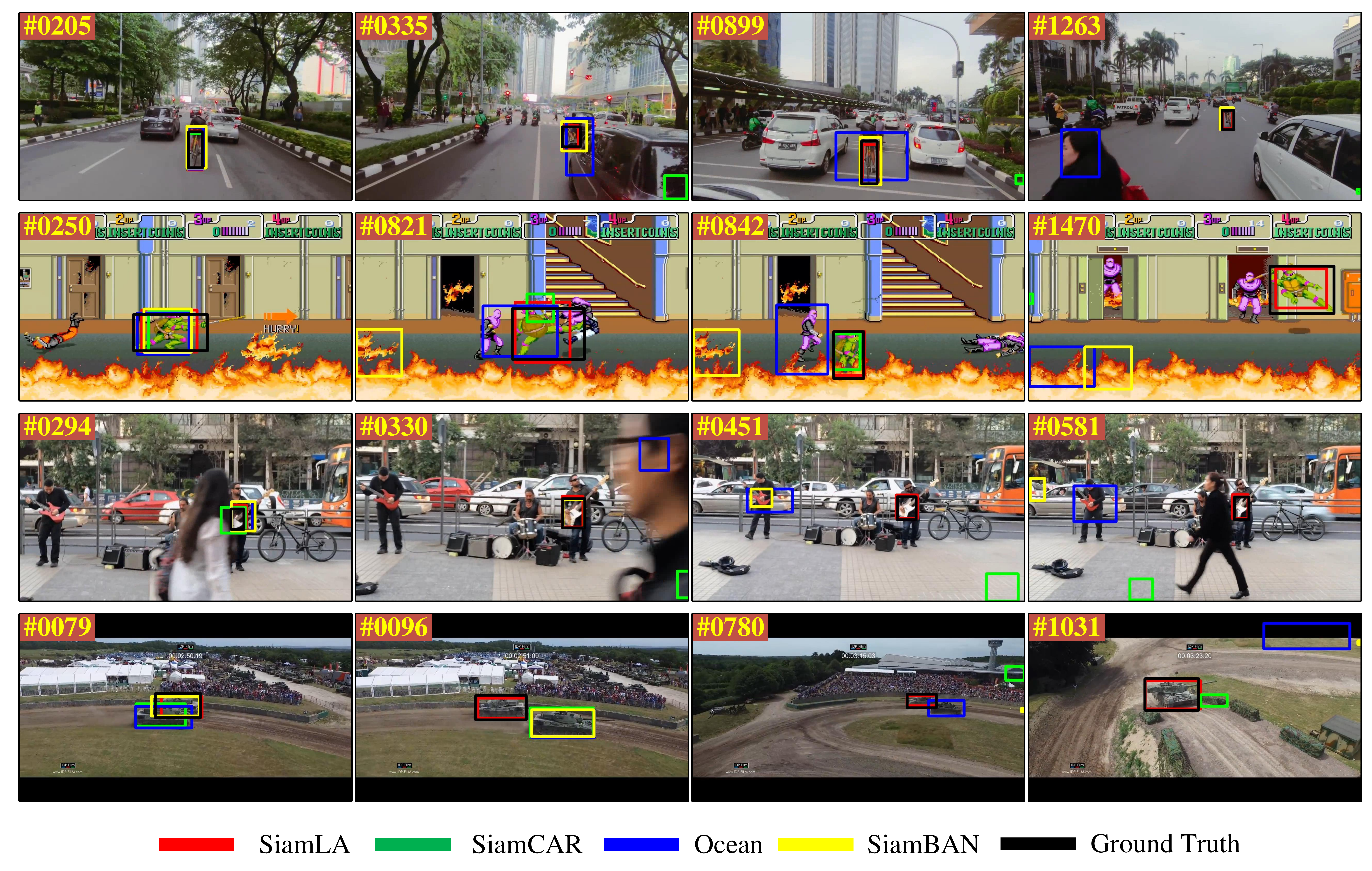}
  \caption{Quantitative comparison of our SiamLA, SiamCAR \cite{siamcar}, Ocean \cite{ocean} and SiamBAN \cite{siamban}. The proposed tracking paradigm, with localization-aware components, robustly predicts more accurate boxes.}
  \label{visual_tracking}
\end{figure}

\textbf{LaSOT: }To further evaluate the proposed tracking paradigm, we conduct experiments on the challenging long-term benchmark LaSOT \cite{lasot}, which consists of 1400 video sequences (280 for test set) with 3.52 million frames. It is labeled with 14 different attributes. We report the results in Table \ref{table-gtl}, the proposed tracker SiamLA gets the competitive performance. Compared with the classic anchor free trackers SiamCAR \cite{siamcar}, Ocean \cite{ocean}, SiamFC++ \cite{siamfc++} and SiamBAN \cite{siamban}, our SiamLA performs better in terms of both accuracy and efficiency, running at real-time speed of 29 fps. Notably, we obtain the best N$\_$Prec. score of 0.652. Moreover, we select some state-of-the-art trackers to report a comparative analysis in 14 different attributes. As shown in Fig. \ref{lasot_attr_suc}, the proposed SiamLA achieves optimal performance for 10 attributes. Compared with the baseline SiamCAR, we obtain performance improvements in 13 attributes ranging from 3\% (rotation) to 8\% (camera motion). Thanks to the proposed separate localization branch, our SiamLA can capture the position of target objects more accurately, and thus leading a great improvement in camera motion attribute. The quantitative results, as shown in Fig. \ref{visual_tracking}, also intuitively present the excellent performance of our tracker.
\begin{figure}[!t]
  \centering
  \includegraphics[width=1.6in]{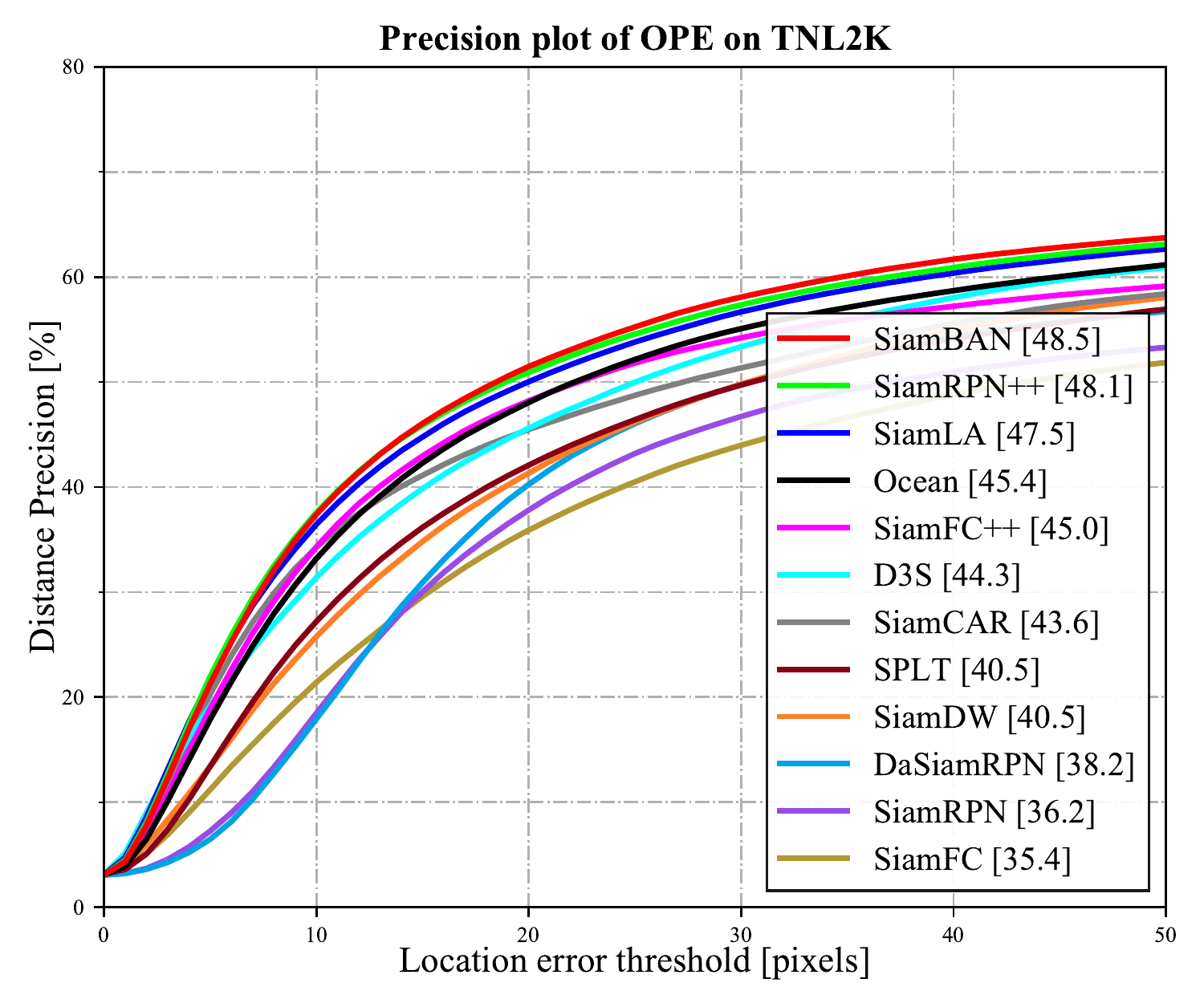}
  \includegraphics[width=1.6in]{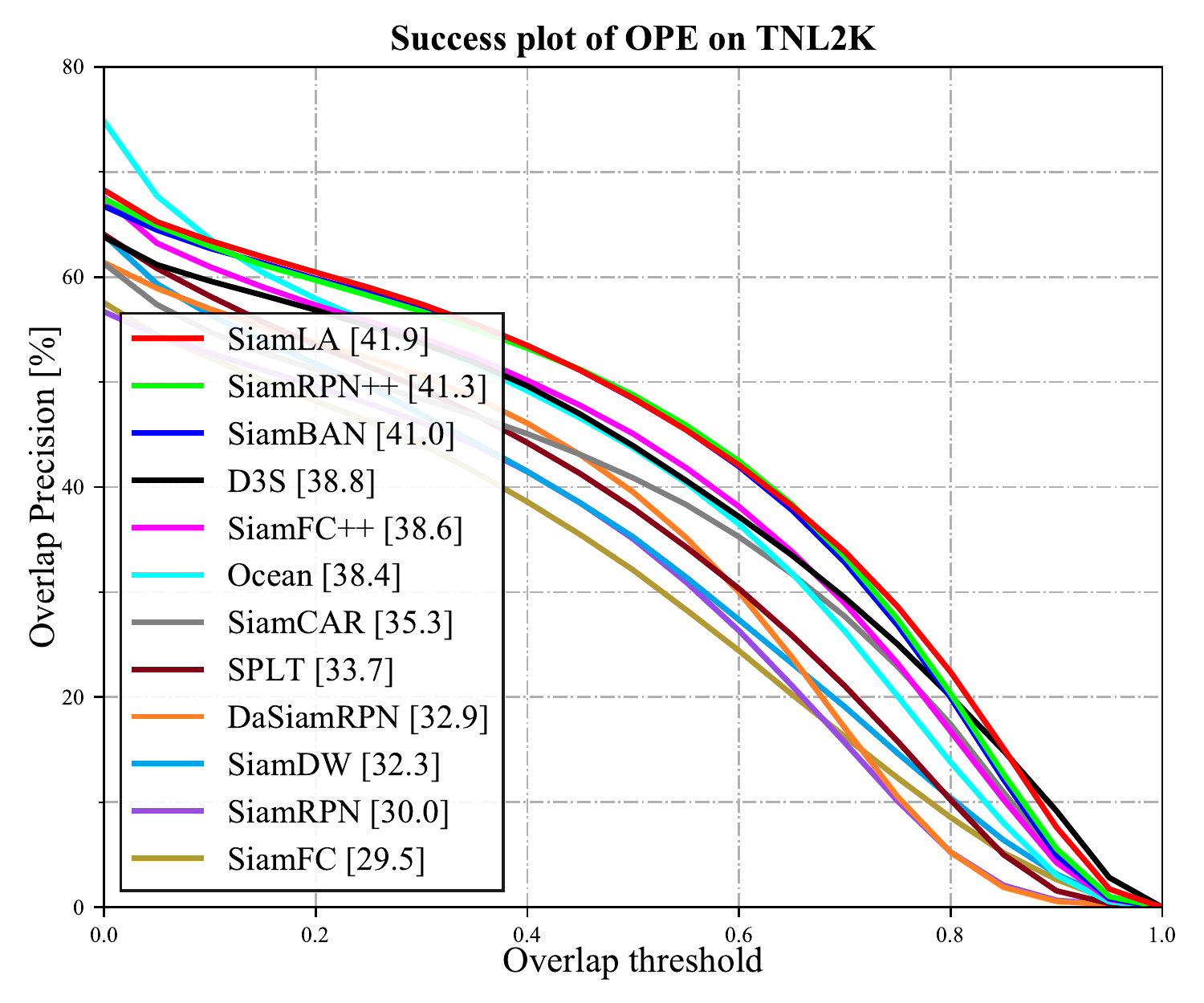}
  \hfill
  \includegraphics[width=1.6in]{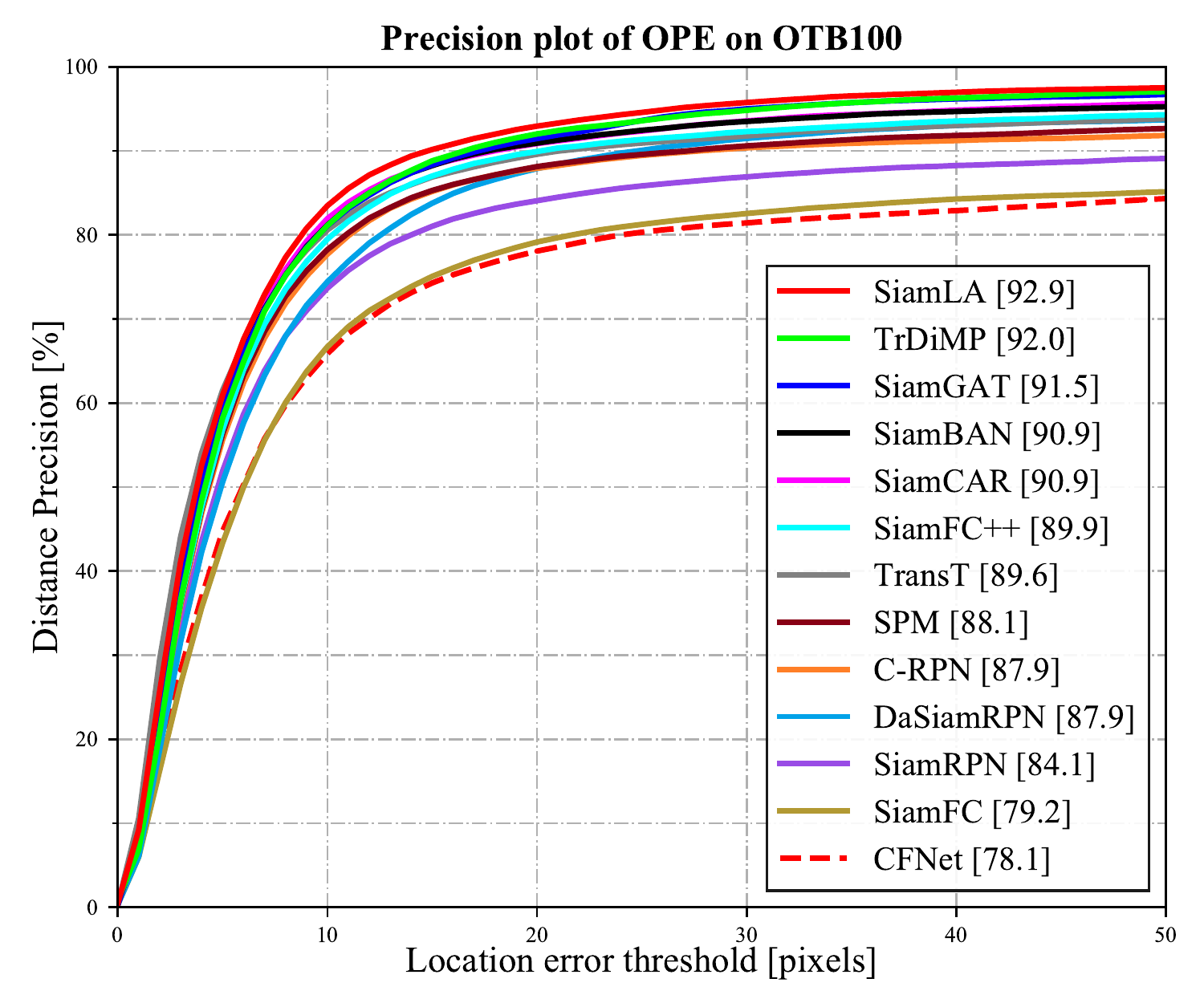}
  \includegraphics[width=1.6in]{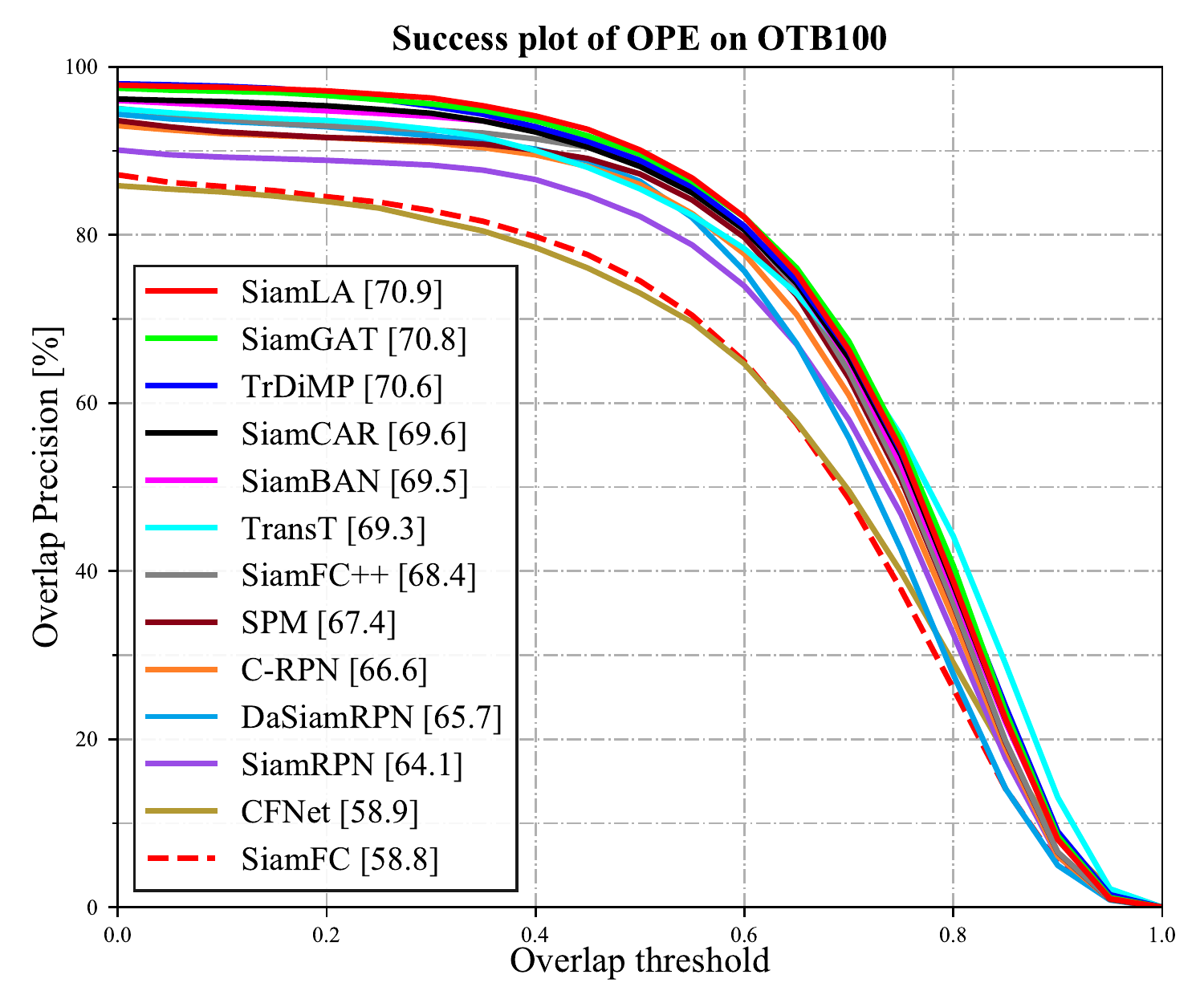}
  \caption{Precision and success plot on the TNL2K \cite{tnl2k} test set and OTB100 \cite{otb2015}, using one-pass evaluation. In the legend, the distance precision at the threshold of 20 pixels (Prec.) in the precision plot and area under the curve (AUC) in the success plot are marked.}
  \label{tnlk2-otb100}
\end{figure}
\begin{figure*}[!b]
  \centering
  \includegraphics[width=7in]{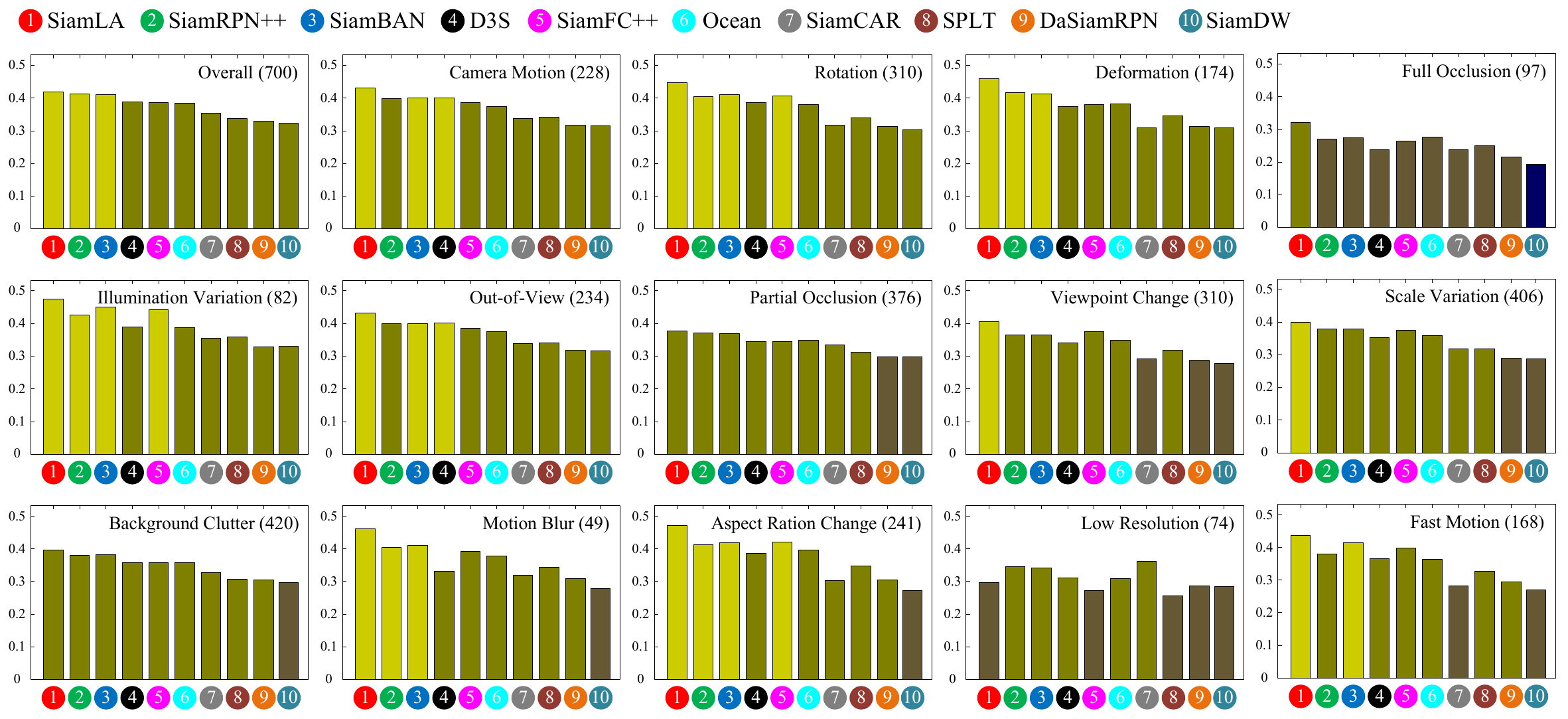}
  \caption{AUC scores of 14 attributes on the TNL2K \cite{tnl2k} test set. The number of video sequences corresponding to each scene is marked, e.g., 700 for overall and 228 for camera motion.}
  \label{tnl2k_attributes}
\end{figure*}

\textbf{TNL2K: }TNL2K \cite{tnl2k} is a recently proposed large-scale benchmark for tracking by natural language and bounding box initialization, with more difficult video sequences. We use the bounding box initialization to evaluate our tracker on 700 video sequence. See Fig. \ref{tnlk2-otb100}, the proposed SiamLA obtains the best AUC score of 0.419, and exceeds 6.3/3.9 points on AUC and Prec., compared to the baseline SiamCAR. In addition, we select the top 10 performance trackers for comparison of 14 attributes. As illustrated in Fig. \ref{tnl2k_attributes}, among the 14 attributes, except low resolution, the SiamLA outperform all the comparison trackers, which proves the proposed tracking paradigm is capable of handling various complex scenes.

\textbf{OTB100: }In addition to the above large-scale benchmarks, we also evaluate the proposed tracker SiamLA on several popular small benchmarks. OTB100 \cite{otb2015} is a one of the most common benchmarks used in tracking community, containing 100 video sequences with 11 types of attributes. As shown in Fig. \ref{tnlk2-otb100}, the SiamLA achieves the advanced performance of 0.709 and 0.929 by the proposed localization-aware components.

\begin{figure}[!t]
  \centering
  \includegraphics[width=3.5in]{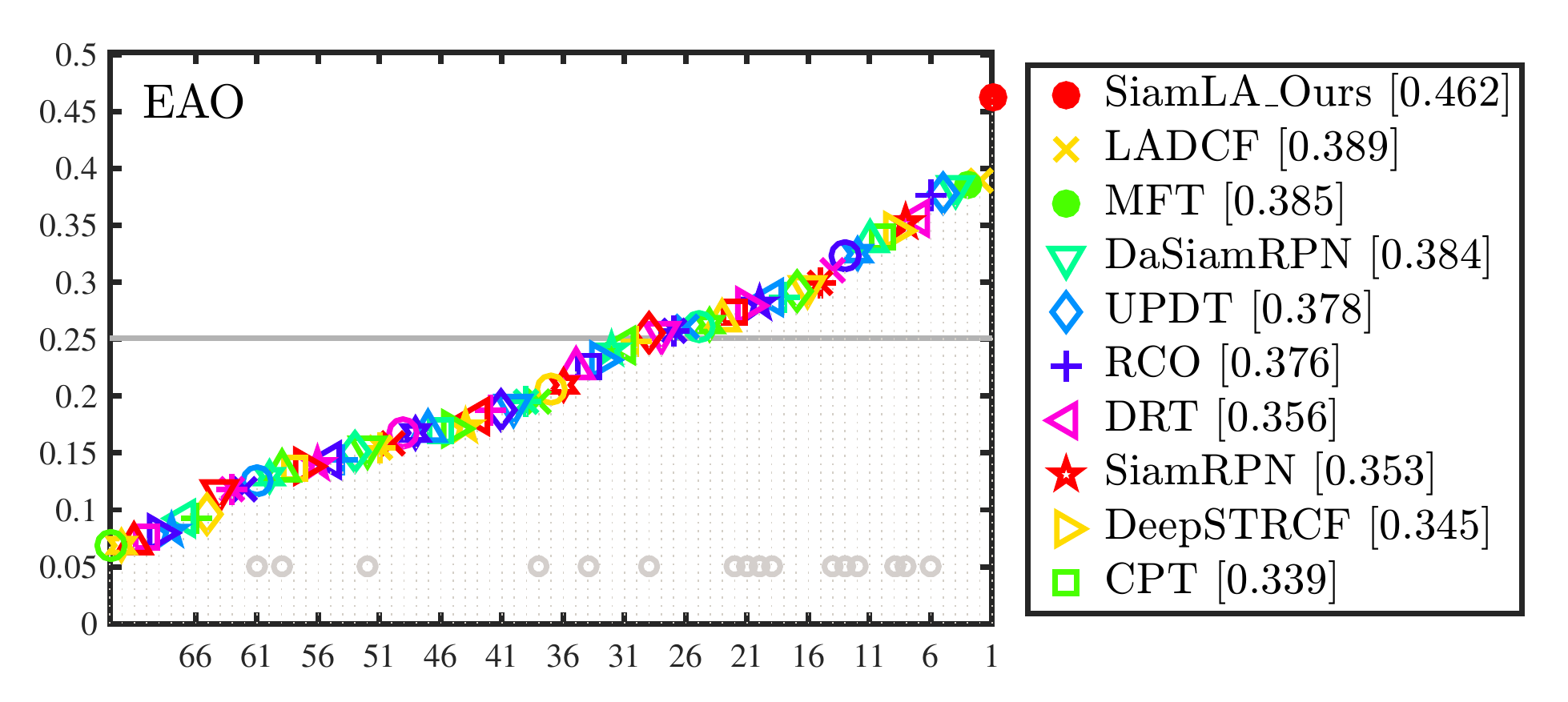}
  \caption{Expected average overlap (EAO) graph with trackers ranked from right to left. Our tracker SiamLA outperforms all the participant trackers on the VOT2018 \cite{vot2018}.}
  \label{vot2018_eaorank}
\end{figure}

\textbf{VOT2018: }VOT2018 \cite{vot2018} is also a small benchmark, which contains a total of 60 video sequences, serving as the benchmark for the seventh visual object tracking challenge. We conduct two sets of experiments to evaluate our tracker SiamLA. First, the state-of-the-comparison results are presented in Table \ref{vot2018}, the SiamLA obtains 0.462, 0.593 and 0.136 scores in terms of EAO, accuracy and robustness, respectively. Then in Fig. \ref{vot2018_eaorank}, we exhibit an EAO ranking, and the proposed tracker SiamLA far outperforms the winner LADCF \cite{ladcf} in VOT2018.
\begin{table}[!h]
  \caption{State-of-the-art Comparison on VOT2018 \cite{vot2018}. The best three results are colored by red, green and blue.}
  \label{vot2018}
  \centering
  \begin{tabular}{c c c c}
    \toprule
    Tracker	&   EAO$\uparrow$ & Accuracy$\uparrow$ & Robustness$\downarrow$  \\
    \midrule
    SiamFC \cite{siamfc}  &   0.188 & 0.503  & 0.585\\	
    DaSiamRPN \cite{siamrpn} &   0.384 & 0.588 & 0.276\\
    Siam R-CNN \cite{siamrcnn} &   0.408 & {\color{green}0.609}  & 0.220\\
    SiamRPN++ \cite{siamrpn++}   & 0.414 & 0.600  & 0.234\\
    \rowcolor{black!10} SiamCAR \cite{siamcar} &   0.423 & 0.574  & 0.197\\
    SiamFC++ \cite{siamfc++} &  0.426 & 0.587  & 0.183\\
    PGNet \cite{pgnet} &   0.447 & {\color{red}0.618} & 0.192\\
    STMTracker \cite{stmtracker} &   0.447 & 0.590  & {\color{blue}0.159}\\
    SiamBAN \cite{siamban}   & 0.452 & 0.597  & 0.178\\
    SiamRCR \cite{siamrcr} &   0.457 & 0.588  & 0.188\\
    TrDiMP \cite{trdimp}&  {\color{blue}0.462} & {\color{blue}0.600}  & {\color{green}0.141}\\
    Ocean \cite{ocean} &  {\color{red}0.467} & 0.598& 0.169\\
    \midrule
    \rowcolor{black!10} SiamLA &   {\color{green}0.462} & 0.593 & {\color{red}0.136} \\	
    \bottomrule
    \end{tabular}
\end{table}

\subsection{Ablation study}
\begin{table*}[!t]
  \caption{Component-wise ablation results on the OTB100 \cite{otb2015} (using AUC metric) and GOT-10k \cite{got10k} (using AO metric) benchmarks.}
  \label{component-wise}
  \centering
  \begin{tabular}{c|ccc|cc|c}
    \toprule 
    Tracking Variation & LADL & LALS & LAFA & OTB100 \cite{otb2015} (AUC) & GOT-10k \cite{got10k} (AO)  &  Speed (fps) \\
    \hline
      SiamCAR &  &  &  & 0.597  & 0.581 & 38 \\
    \hline 
    \multirow{5}{*}{SiamLA}
       & \checkmark &  &   &  0.621 & 0.594
    & \multirow{3}{*}{38} \\
       &  & \checkmark &  & 0.573  & 0.571 \\
       & \checkmark & \checkmark &   & 0.632 & 0.602 \\
    \cline{2-7}
       &  &  & \checkmark & 0.626  & 0.599
       & \multirow{2}{*}{29} \\
       & \checkmark & \checkmark & \checkmark & 0.649 & 0.619\\
    \bottomrule
  \end{tabular}
\end{table*}
\textbf{Component-wise analysis: }To verify the effectiveness of the proposed localization-aware components (i.e., LADL, LALS and LAFA), and investigate their contribution to the proposed tracker SiamLA, we conduct an ablation experiment on the OTB100 \cite{otb2015} and GOT-10k \cite{got10k} benchmarks. Moreover, we also compare the running speed of the tracking variations with different components. To be fair, we train the tracking variations using GOT-10k training set, and evaluate them with the same hyper-parameters in testing. We report the ablation results in Table \ref{component-wise}. The proposed LADL enables collaborative optimization between the classification and regression, making the accurate prediction boxes are predicted as high target confidence scores. Thus, 2.4\% and 1.3\% improvements in AUC and AO are obtained, respectively. Besides, no speed loss is involved, because the LADL component has no additional time-space complexity in testing. For the LALS component, it is proposed to define the positive labels for boundary points. However, when using this strategy without the LADL, the performance drops. This is due to that the additional positive samples distract the model. But when using them jointly, the performance is further enhanced. The proposed separate localization branch (represented by LVFA), delivers the location quality scores to modify the classification scores, also leading to significant performance advantages, in which 2.9\% and 1.8\% in AUC and AO scores. Moreover, despite slower 9 fps than the baseline, it is still exceeds the real-time speed (24 fps). With all these proposed components, our tracker gets the best performance of 0.649 and 0.619 in terms of AUC and AO scores, which well confirms the effectiveness of the proposed localization-aware components in this work.
\begin{figure}[!h]
  \centering
  \includegraphics[width=1.6in]{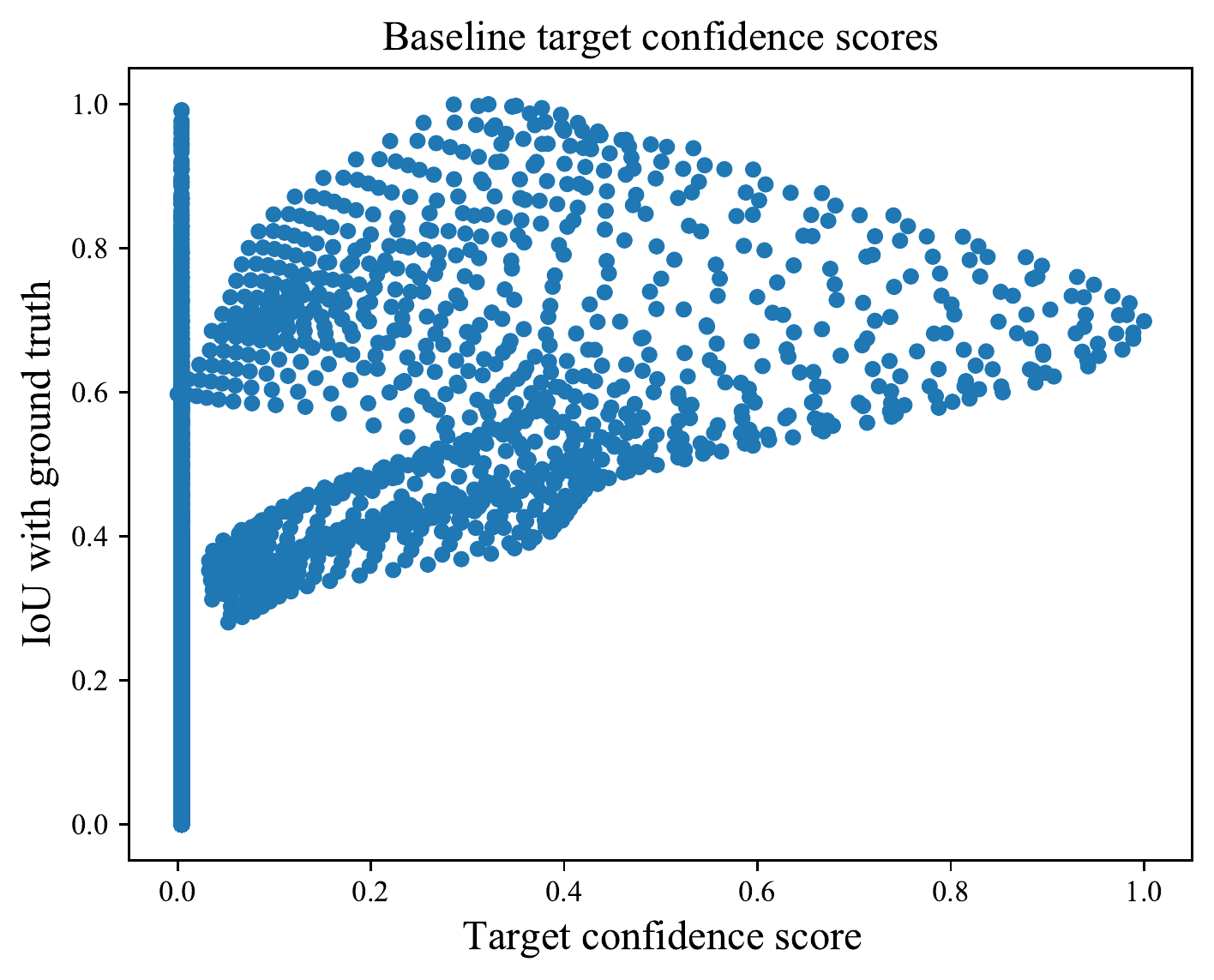}
  \includegraphics[width=1.6in]{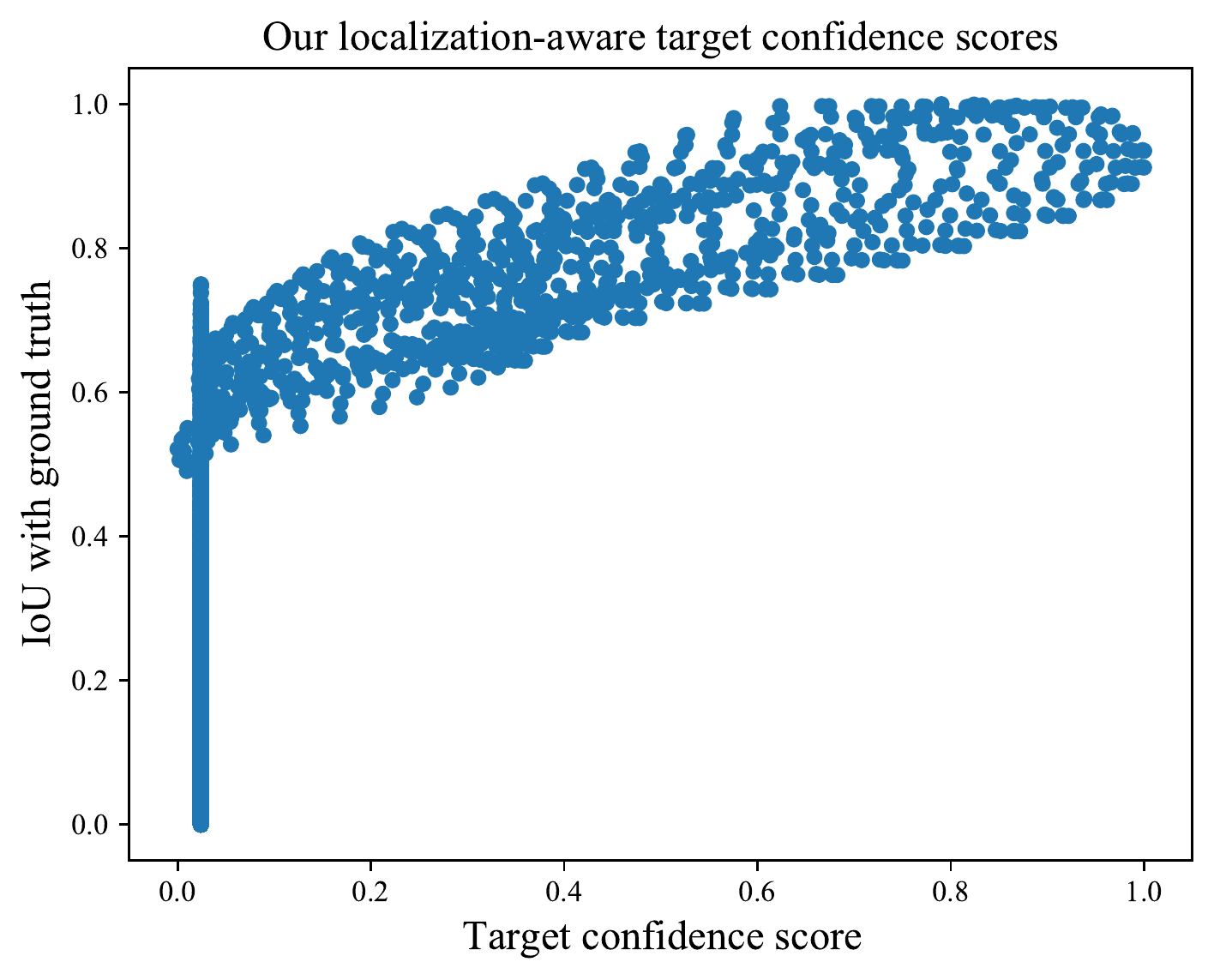}
  \caption{The correlation between IoUs and the target confidence scores. On the left: \textbf{baseline}, the target confidence scores are the classification scores multiplied by the Center-ness scores. On the right: \textbf{Ours}, the target confidence scores are the classification scores multiplied by the location quality scores.}
  \label{target confidence score}
\end{figure}
\begin{figure*}[!t]
  \centering
  \includegraphics[width=5in]{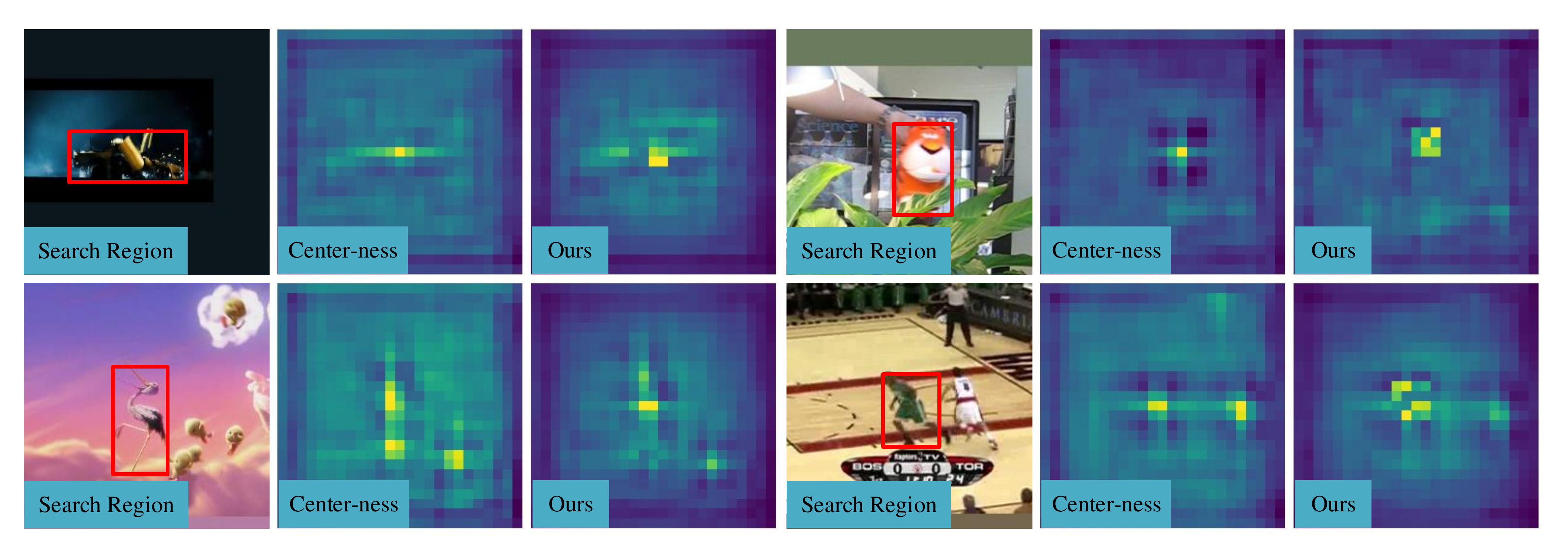}
  \caption{The localization quality assessments scores of the Center-ness and the proposed separate branch (represented by Ours). Our component and Center-ness show a different way of assessment, and the distractors are effectively suppressed by the separate branch.}
  \label{centernessVS.ours}
\end{figure*}

To intuitively present how the proposed tracking paradigm alleviate the task misalignment, we perform some visualization comparison experiments. As already stated in testing details, the selection of the predicted box depends on the target confidence scores. Therefore, a positive correlation between the IoUs and the target confidence scores implies perfect task alignment. As shown in Fig. \ref{target confidence score}, our tracker has a higher positive correlation compared to the baseline, indicating that the proposed localization-aware components effectively alleviate the task misalignment. In addition, we also visualize the location quality scores generated by the Center-ness and the proposed separate branch. As illustrated in Fig. \ref{centernessVS.ours}, the top row clearly presents that the Center-ness predicts the potential target center, not the location accuracy. Differently, our separate branch predicts more points, making accurate prediction boxes predicted as high scores. Besides, the bottom shows that the distractors are suppressed effectively with the proposed branch, while the Center-ness fails. This is due to that we incorporate rich localization-aware information by the LAFA module, and sample some negative samples to train it.

\textbf{Localization-aware dynamic label: }As illustrated in Equation \ref{eq4}, we map the linearly varying $IoU_{i,j}$ to the nonlinear $\hat{IoU_{i,j}}$, making the classification easier to be optimized, and thus leading to a performance increase. To prove this, we conduct a comparison experiment of the linear IoU and nonlinear IoU. The results presented in table \ref{ladl}, demonstrating the effectiveness of the nonlinear design in Equation \ref{eq4}.
\begin{table}[!h]
  \caption{Ablation results of the proposed \textit{localization-aware dynamic label} on the OTB100 \cite{otb2015} (using AUC metric) and GOT-10k \cite{got10k} (using AO metric) benchmarks.}
  \label{ladl}
  \centering
  \begin{tabular}{c|cc}
    \toprule
      Tracking Variation & OTB100 (AUC) &  GOT-10k (AO)  \\
    \hline
      Linear $IoU_{i,j}$ & 0.613  &  0.591 \\
      Nonlinear $\hat{IoU_{i,j}}$ & 0.621  & 0.594 \\
    \bottomrule
  \end{tabular}
\end{table}

\textbf{Localization-aware label smoothing: }We also exhibit a comparison experiment of the parameter $\lambda$ in Equation \ref{eq6}, a significant hyper-parameter in the proposed LALS strategy. A larger $\lambda$ implies more high-quality samples will be included, however, more distractors is introduced at the same time, preventing the model from selecting the most accurate prediction box. As shown in Table \ref{lals}, when $\lambda=0.2$, the tracker performs best, obtaining 0.632 and 0.602 scores in terms of AUC and AO scores. Therefore, in all experiments, we set $\lambda$ to 0.2.
\begin{table}[!h]
  \caption{Ablation results of the proposed \textit{localization-aware label smoothing} module on the OTB100 \cite{otb2015} (using AUC metric) and GOT-10k \cite{got10k} (using AO metric) benchmarks.}
  \label{lals}
  \centering
  \begin{tabular}{c|cc}
    \toprule
      $\lambda$  $$& OTB100 (AUC) &  GOT-10k (AO)  \\
    \hline
      0.4 &  0.604 &  0.583\\
      0.3 &  0.624 &  0.594\\
      0.2 &  0.632 &  0.602\\
      0.1 &  0.619 &  0.591\\
    \bottomrule
  \end{tabular}
\end{table}
\begin{table*}[!t]
  \caption{Ablation results of the proposed \textit{localization-aware feature aggregation} module on the OTB100 \cite{otb2015} (using AUC metric) and GOT-10k \cite{got10k} (using AO metric) benchmarks.}
  \label{lafa}
  \centering
  \begin{tabular}{c|cc}
    \toprule
      Tracking Variation & OTB100 (AUC) &  GOT-10k (AO)  \\
    \hline
      LAFA & 0.626  &  0.599\\
      LAFA w/o localization-aware non-local block & 0.617  & 0.594 \\
      LAFA w/o feature aggregation block & 0.611  &  0.591 \\
    \bottomrule
  \end{tabular}
\end{table*}

\textbf{Localization-aware feature aggregation: }According to the Table \ref{component-wise}, the ablation experiment has demonstrated the effectiveness of the separate localization branch. As its core module, the LAFA plays a significant role in guiding the more accurate location quality scores. To further investigate the localization-aware non-local block and feature aggregation block existing in the LAFA module, another ablation experiment is performed. The results are presented in Table \ref{lafa}. Without localization-aware non-local block, 0.9 and 0.5 points of AUC and AO scores drop. While without feature aggregation block, the AUC and AO scores drop by 1.5 and 0.8 points. 

\subsection{Stability analysis}
\begin{figure}[!h]
  \centering
  \includegraphics[width=2in]{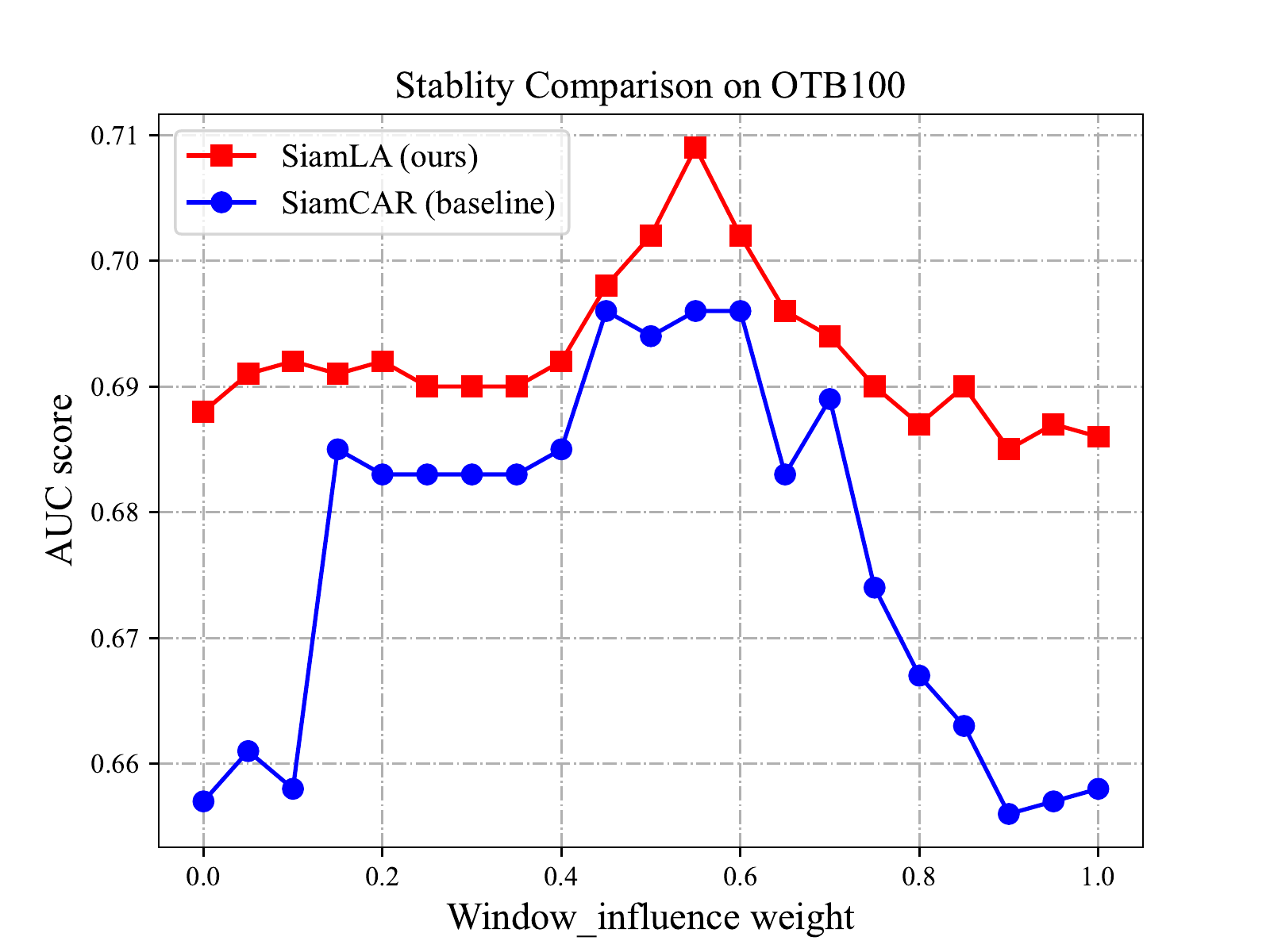}
  \caption{Comparison of the AUC scores with different window$\_$influence on OTB100 \cite{otb2015}.}
  \label{stable}
\end{figure}

In fact, the Siamese paradigm is tend to insufficiently stable, making them difficult to apply to real-world scenes. This deficiency stems from a hyper-parameters, window$\_$influence, as shown in https://github.com/STVIR/pysot, also can be seen in Equation \ref{eq11}. We conduct an extra experiment to demonstrate the stability of the proposed tracking paradigm. Fig. \ref{stable} presents the comparison results, we clearly observe that our tracker SiamLA is more stable than the baseline SiamCAR. Especially without post-processing in testing (i.e., window$\_$influence = 0), our tracker still has competitive performance (0.687 AUC score, suggesting that the proposed tracking paradigm is superior at mining the optimal prediction box, making it more potential to real-world applications. 

\section{Conclusion}
This paper presents a novel Siamese tracking paradigm, resolving the longstanding unsolved misalignment problem between classification and regression. We first analyze how the task misalignment problem manifests and why it arises. To overcome it, we propose a series of localization-aware components, i.e., \textit{localization-aware dynamic label} (LADL), \textit{localization-aware label smoothing} (LALS) and \textit{localization-aware feature aggregation} (LAFA) to synchronize the classification and regression. Our tracker SiamLA is evaluated on six challenging benchmarks. The results show that the SiamLA achieves competitive performance compared to the state-of-the-art trackers, which demonstrate the effectiveness of the proposed tracking paradigm. We also conduct some visualization experiments to clearly present how the paradigm alleviate the task misalignment. Moreover, a stability analysis reveals that our tracking paradigm is relatively stable.

In fact, specific tasks require specific forms of feature representation. However, in the mainstream algorithms, the classification and regression usually are built with same structure, which contradicts the specific forms of feature representation. Therefore, future work will focus on this issue.




\begin{thebibliography}{1}

  \bibitem{app}
  Z. Sun, J. Chen, L. Chao, W. Ruan and M. Mukherjee, ``A Survey of Multiple Pedestrian Tracking Based on Tracking-by-Detection Framework,'' \textit{IEEE Trans. Circuits and Syst. Video Techn.}, vol. 31, no. 5, pp. 1819-1833, May. 2021.

  \bibitem{things}
  J. Zhang and D. Tao, ``Empowering Things With Intelligence: A Survey of the Progress, Challenges, and Opportunities in Artificial Intelligence of Things," \textit{IEEE Internet Things J.}, vol. 8, no. 10, pp. 7789-7817, 2021.

  \bibitem{survey}
  S. M. Marvasti-Zadeh, L. Cheng, H. Ghanei-Yakhdan and S. Kasaei, ``Deep Learning for Visual Tracking: A Comprehensive Survey," \textit{IEEE Trans. Intell. Transp. Syst.}, pp. 1-26, 2021.

  \bibitem{vot2020}
  M. Kristan, \textit{et al}., ``The eighth visual object tracking VOT2020 challenge results," in \textit{Proc. Eur. Conf. Comput. Vis. (ECCV)}, Oct. 2020, pp. 547-601.

  \bibitem{vot2021}
  M. Kristan, \textit{et al}., ``The ninth visual object tracking VOT2021 challenge results," in \textit{proc. IEEE/CVF Int. Conf. Comput. Vis. (ICCV)}, Oct. 2021, pp. 2711-2738.

  \bibitem{otb2013}
  Y. Wu, J. Lim and M. Yang, ``Online Object Tracking: A Benchmark," in \textit{proc. IEEE Conf. Comput. Vis. Pattern Recognit. (CVPR)}, Jun. 2013, pp. 2411-2418.
  
  \bibitem{uav123}
  M. Mueller, N. Smith and B. Ghanem, ``A benchmark and simulator for uav tracking," in \textit{Proc. Eur. Conf. Comput. Vis. (ECCV)}, Oct. 2016, pp. 445-461.
  
  \bibitem{nfs}
  H. K. Galoogahi, A. Fagg, C. Huang, D. Ramanan and S. Lucey, ``Need for Speed: A Benchmark for Higher Frame Rate Object Tracking," in \textit{proc. IEEE Int. Conf. Comput. Vis. (ICCV)}, Oct. 2017, pp. 1134-1143.
  
  \bibitem{siamfc}
  L. Bertinetto \textit{et al}., ``Fully-convolutional siamese networks for object tracking," in \textit{Proc. Eur. Conf. Comput. Vis. (ECCV)}, Oct. 2016, pp. 850-865.
  
  \bibitem{siamrpn}
  B. Li, J. Yan, W. Wu, Z. Zhu and X. Hu, ``High Performance Visual Tracking with Siamese Region Proposal Network," in \textit{proc. IEEE/CVF Conf. Comput. Vis. Pattern Recognit. (CVPR)}, Jun. 2018, pp. 8971-8980.
  
  \bibitem{siamcar}
  D. Guo, J. Wang, Y. Cui, Z. Wang and S. Chen, ``SiamCAR: Siamese Fully Convolutional Classification and Regression for Visual Tracking," in \textit{proc. IEEE Conf. Comput. Vis. Pattern Recognit. (CVPR)}, Jun. 2020, pp. 6268-6276.
  
  \bibitem{ocean}
  Z. Zhang, \textit{et al}, ``Ocean: Object-aware anchor-free tracking," in \textit{Proc. Eur. Conf. Comput. Vis. (ECCV)}, Aug. 2020, pp. 771-787.

  \bibitem{siamfc++}
  Y. Xu, Z. Wang, Z. Li, Y. Yuan and G. Yu, ``SiamFC++: Towards robust and accurate visual tracking with target estimation guidelines," in \textit{Proc. AAAI Conf. Artif. Intell. (AAAI)}, Feb. 2020, pp. 12549-12556.
  
  \bibitem{siamban}
  Z. Chen, B. Zhong, G. Li, S. Zhang and R. Ji, ``Siamese Box Adaptive Network for Visual Tracking," in \textit{proc. IEEE/CVF Conf. Comput. Vis. Pattern Recognit. (CVPR)}, Jun. 2020, pp. 6667-6676.
  
  \bibitem{deeplearning1}
  Y. Xu, \textit{et al}, ``Vitae: Vision transformer advanced by exploring intrinsic inductive bias," in \textit{Proc. Adv. Neural Inf. Process. Syst. (NIPS)}, 2021, 34.
  
  \bibitem{deeplearning2}
  Q. Zhang, \textit{et al}, ``Vitaev2: Vision transformer advanced by exploring inductive bias for image recognition and beyond," 2022, \textit{arxiv:2202.10108}. [Online]. Available: https://arxiv.org/abs/arxiv:2202.10108

  \bibitem{dasiamrpn}
  Z. Zhu, Q. Wang, B. Li, W. Wu, J. Yan and W. Hu, ``Distractor-aware siamese networks for visual object tracking," in \textit{Proc. Eur. Conf. Comput. Vis. (ECCV)}, Sep. 2018, pp. 101-117.
  
  \bibitem{siamrpn++}
  B. Li, W. Wu, Q. Wang, F. Zhang, J. Xing and J. Yan, ``SiamRPN++: Evolution of Siamese Visual Tracking With Very Deep Networks," in \textit{proc. IEEE/CVF Conf. Comput. Vis. Pattern Recognit. (CVPR)}, Jun. 2019, pp. 4277-4286.
  
  \bibitem{cfnet}
  J. Valmadre, L. Bertinetto, \textit{et al}, ``End-to-end representation learning for correlation filter based tracking," in \textit{proc. IEEE/CVF Conf. Comput. Vis. Pattern Recognit. (CVPR)}, Jun. 2017, pp. 2805-2813.

  \bibitem{sasiam}
  A. He, C. Luo, X. Tian, and W. Zeng, ``A twofold siamese network for real-time object tracking," in \textit{proc. IEEE/CVF Conf. Comput. Vis. Pattern Recognit. (CVPR)}, Jun. 2018, pp. 4834-4843.

  \bibitem{structsiam}
  Y. Zhang, L. Wang, J. Qi, \textit{et al}, ``Structured siamese network for real-time visual tracking," in \textit{Proc. Eur. Conf. Comput. Vis. (ECCV)}, Sep. 2018 2018: 351-366.

  \bibitem{siamdw}
  Z. Zhang, \textit{et al}, ``Deeper and wider siamese networks for real-time visual tracking," in \textit{proc. IEEE/CVF Conf. Comput. Vis. Pattern Recognit. (CVPR)}, Jun. 2019, pp. 4591-4600.

  \bibitem{dp}
  Y. Zha, T. Ku, Y. Li and P. Zhang, ``Deep Position-Sensitive Tracking," \textit{IEEE Trans. Multimedia.}, vol. 22, no. 1, pp. 96-107, 2020.

  \bibitem{fastrcnn}
  R. Girshick, \textit{et al}, ``Fast r-cnn," in \textit{proc. IEEE Int. Conf. Comput. Vis. (ICCV)}, Oct. 2015, pp. 1440-1448.

  \bibitem{alxenet}
  A. Krizhevsky, I. Sutskever and G. Hinton, ``Imagenet classification with deep convolutional neural networks", in \textit{Proc. Adv. Neural Inf. Process. Syst. (NIPS)}, Dec. 2012, pp. 25.

  \bibitem{resnet}
  K. He, X. Zhang, S. Ren and J. Sun, ``Deep Residual Learning for Image Recognition," in \textit{proc. IEEE Conf. Comput. Vis. Pattern Recognit. (CVPR)}, Jun. 2016, pp. 770-778.

  \bibitem{pgnet}
  B. Liao, \textit{et al}, ``Pg-net: Pixel to global matching network for visual tracking," in \textit{Proc. Eur. Conf. Comput. Vis. (ECCV)}, Aug. 2020, pp. 429-444.

  \bibitem{siamacm}
  W. Han, X. Dong, \textit{et al}, ``Learning to fuse asymmetric feature maps in siamese trackers," in \textit{proc. IEEE/CVF Conf. Comput. Vis. Pattern Recognit. (CVPR)}, Jun. 2021, pp. 16570-16580.

  \bibitem{siamgat}
  D. Guo, Y. Shao, Y. Cui, Z. Wang, L. Zhang and C. Shen, ``Graph Attention Tracking," in \textit{proc. IEEE/CVF Conf. Comput. Vis. Pattern Recognit. (CVPR)}, Jun. 2021, pp. 9538-9547.
  
  \bibitem{transt}
  X. Chen, B. Yan, J. Zhu, D. Wang, X. Yang and H. Lu, ``Transformer Tracking," in \textit{proc. IEEE/CVF Conf. Comput. Vis. Pattern Recognit. (CVPR)}, Jun. 2021, pp. 8122-8131.

  \bibitem{learntomatch}
  Z. Zhang, Y. Liu, X. Wang, \textit{et al}, ``Learn to match: Automatic matching network design for visual tracking," in \textit{proc. IEEE/CVF Int. Conf. Comput. Vis. (ICCV)}, Oct. 2021, pp. 13339-13348.

  \bibitem{ttdimp}
  J. Nie, H. Wu, Z. He, M. Gao and Z. Dong, ``Spreading Fine-grained Prior Knowledge for Accurate Tracking," \textit{IEEE Trans. Circuits and Syst. Video Techn.}, early access, 2022, doi: 10.1109/TCSVT.2022.3162599.

  \bibitem{multi-level}
  Q. Liu, X. Li, Z. He, N. Fan, D. Yuan and H. Wang, ``Learning Deep Multi-Level Similarity for Thermal Infrared Object Tracking," \textit{IEEE Trans. Multimedia.}, vol. 23, pp. 2114-2126, 2021.

  \bibitem{bt}
  M. Li, L. Peng, T. Wu and Z. Peng, ``A Bottom-Up and Top-Down Integration Framework for Online Object Tracking," \textit{IEEE Trans. Multimedia.}, vol. 23, pp. 105-119, 2021.

  \bibitem{siamcorners}
  K. Yang, \textit{et al}, ``SiamCorners: Siamese Corner Networks for Visual Tracking," \textit{IEEE Trans. Multimedia.}, vol. 24, pp. 1956-1967, 2022.

  \bibitem{cat}
  S. Zhang, X. Zhao and L. Fang, ``CAT: Corner Aided Tracking With Deep Regression Network,'' \textit{IEEE Trans. Multimedia.}, vol. 23, pp. 859-870, 2021.

  \bibitem{spm}
  G. Wang, C. Luo, Z. Xiong and W. Zeng, ``SPM-Tracker: Series-Parallel Matching for Real-Time Visual Object Tracking," in \textit{proc. IEEE/CVF Conf. Comput. Vis. Pattern Recognit. (CVPR)}, Jun. 2019, pp. 3638-3647.
  
  \bibitem{c-rpn}
  H. Fan and H. Ling, ``Siamese Cascaded Region Proposal Networks for Real-Time Visual Tracking," in \textit{proc. IEEE/CVF Conf. Comput. Vis. Pattern Recognit. (CVPR)}, Jun. 2019, pp. 7944-7953.
  
  \bibitem{multi-stage}
  G. Han, J. Su, Y. Liu, Y. Zhao and S. Kwong, ``Multi-Stage Visual Tracking with Siamese Anchor-Free Proposal Network," \textit{IEEE Trans. Multimedia.}, doi: 10.1109/TMM.2021.3127357.

  \bibitem{siamrcr}
  J. Peng \textit{et al}, ``SiamRCR: Reciprocal Classification and Regression for Visual Object Tracking," 2021, \textit{arxiv:2105.11237}. [Online]. Available: https://arxiv.org/abs/arxiv:2105.11237
  
  \bibitem{fcos}
  Z. Tian, C. Shen, H. Chen, \textit{et al}, ``Fcos: Fully convolutional one-stage object detection," in \textit{proc. IEEE/CVF Conf. Comput. Vis. Pattern Recognit. (CVPR)}, Jun. 2019, pp. 9627-9636.

  \bibitem{trdimp}
  N. Wang, W. Zhou, J. Wang and H. Li, ``Transformer Meets Tracker: Exploiting Temporal Context for Robust Visual Tracking," in \textit{proc. IEEE/CVF Conf. Comput. Vis. Pattern Recognit. (CVPR)}, Jun. 2021, pp. 1571-1580.
  
  \bibitem{tomp}
  C. Mayer, \textit{et al}, ``Transforming Model Prediction for Tracking," 2022, \textit{arxiv:22.3.11192}. [Online]. Available: https://arxiv.org/abs/arxiv:2203.11192

  \bibitem{imagenet}
  O. Russakovsky, \textit{et al}, ``Imagenet large scale visual recognition challenge," in \textit{Int. Journal of Comput. Vis.}, vol. 115, no. 3, pp. 211-252, 2015.
  
  \bibitem{ytbb}
  E. Real, J. Shlens, S. Mazzocchi, X. Pan, V. Vanhoucke, ``Youtube-bounding boxes: A large high-precision human-annotated dataset for object detection in video," in \textit{proc. IEEE/CVF Conf. Comput. Vis. Pattern Recognit. (CVPR)}, Jun. 2017, pp. 5296-53050.

  \bibitem{got10k}
  L. Huang, X. Zhao and K. Huang, ``GOT-10k: A Large High-Diversity Benchmark for Generic Object Tracking in the Wild," \textit{IEEE Trans. Intell. Transp. Syst.}, vol. 43, no. 5, pp. 1562-1577, 1 May 2021.

  \bibitem{coco}
  L. TY \textit{et al}., ``Microsoft coco: Common objects in context," in \textit{Proc. Eur. Conf. Comput. Vis. (ECCV)}, Sep. 2014, pp. 740-755.
  
  \bibitem{trackingnet}
  M. Müller, A. Bibi, S. Giancola, S. Al-Subaihi and B. Ghanem, ``Trackingnet: A large-scale dataset and benchmark for object tracking in the wild," in \textit{Proc. Eur. Conf. Comput. Vis. (ECCV)}, Sep. 2018, pp. 300-317.

  \bibitem{lasot}
  F. Heng \textit{et al}., ``LaSOT: A High-quality Benchmark for Large-scale Single Object Tracking," in \textit{proc. IEEE Conf. Comput. Vis. Pattern Recognit. (CVPR)}, Jun. 2019, pp. 5369-5378.

  \bibitem{tnl2k}
  X. Wang, X. Shu, Z. Zhang, \textit{et al}, ``Towards more flexible and accurate object tracking with natural language: Algorithms and benchmark," in \textit{proc. IEEE/CVF Conf. Comput. Vis. Pattern Recognit. (CVPR)}, Jun. 2021, pp. 13763-13773.
  
  \bibitem{otb2015}
  Y. Wu, J. Lim and M. Yang, ``Object Tracking Benchmark,"  \textit{IEEE Trans. Pattern Anal. Mach. Intell.}, vol. 37, no. 9, pp. 1834-1848, Sep. 2015.

  \bibitem{vot2018}
  M. Kristan, \textit{et al}, ``The sixth visual object tracking VOT2018 challenge results," in \textit{proc. IEEE Int. Conf. Comput. Vis. Workshops (ICCVW)}, Sep. 2018, pp. 3-53.

  \bibitem{dsiam}
  Q. Guo, \textit{et al}, ``Learning dynamic siamese network for visual object tracking," in \textit{proc. IEEE Int. Conf. Comput. Vis. (ICCV)}, Oct. 2017, pp. 1763-1771.
  
  \bibitem{siamltr}
  T. Feng and Q. Ling, ``Learning to Rank Proposals for Siamese Visual Tracking,'' \textit{IEEE Trans. Image Process.},vol. 30, pp. 3311-3320, 2021.

  \bibitem{splt}
  B. Yao, H. Zhao, \textit{et al}, ``Skimming-Perusal Tracking: A Framework for Real-Time and Robust Long-Term Tracking," in \textit{proc. IEEE Int. Conf. Comput. Vis. (ICCV)}, Oct. 2019, pp. 2385-2393.

  \bibitem{d3s}
  A. Lukezic, J. Matas and M. Kristan, ``D3S-A Discriminative Single Shot Segmentation Tracker," in \textit{proc. IEEE Conf. Comput. Vis. Pattern Recognit. (CVPR)}, Jun. 2020, pp. 7133-7142.

  \bibitem{roam}
  T. Yang \textit{et al}, ``ROAM: Recurrently Optimizing Tracking Model," in \textit{proc. IEEE Conf. Comput. Vis. Pattern Recognit. (CVPR)}, Jun. 2020, pp. 6718-6727.

  \bibitem{sbtlight}
  F. Xie \textit{et al}, ``Correlation-Aware Deep Tracking," 2022, \textit{arxiv:2203.01666}. [Online]. Available: https://arxiv.org/abs/arxiv:2203.01666

  \bibitem{siamrcnn}
  P. Voigtlaender, J. Luiten, P. H. Torr, and B. Leibe, ``SiamR-CNN: Visual Tracking by Re-Detection," in \textit{proc. IEEE Conf. Comput. Vis. Pattern Recognit. (CVPR)}, Jun. 2020, pp. 6578-6588.

  \bibitem{stmtracker}
  Z. Fu, Q. Liu, Z. Fu, Y. Wang, ``STMTrack: Template-free Visual Tracking with Space-time Memory Networks," in \textit{proc. IEEE/CVF Conf. Comput. Vis. Pattern Recognit. (CVPR)}, Jun. 2021, pp. 13774-13783.

  \bibitem{ladcf}
  T. Xu, Z. -H. Feng, X. -J. Wu and J. Kittler, ``Learning Adaptive Discriminative Correlation Filters via Temporal Consistency Preserving Spatial Feature Selection for Robust Visual Object Tracking," \textit{IEEE Trans. Image Process.}, vol. 28, no. 11, pp. 5596-5609, Nov. 2019.

\end{thebibliography}
\end{document}